%% file: main.tex
\documentclass[10pt,twocolumn,letterpaper]{article}

\usepackage{cvpr}              % To produce the CAMERA-READY version
\input{preamble}
\definecolor{cvprblue}{rgb}{0.21,0.49,0.74}
\usepackage[pagebackref,breaklinks,colorlinks,allcolors=cvprblue]{hyperref}
\usepackage{multirow}
\usepackage[accsupp]{axessibility}  % Improves PDF readability for those with disabilities.

\definecolor{yao}{rgb}{0.0, 0.5, 1.0} % azure
\newcommand{\comment}[1]{\ignorespaces}

\def\paperID{31866} % *** Enter the Paper ID here
\def\confName{CVPR}
\def\confYear{2026}

\title{HCL-FF: Hierarchical and Contrastive Learning for Forward-Forward Algorithm}

\author{
  Jie-En Yao\qquad
  Hong-En Chen\qquad
  C.-C. Jay Kuo\\
  University of Southern California\\
  {\tt\small \{jieenyao,hongench,jckuo\}@usc.edu}
}

\begin{document}
\maketitle
\input{sec/0_abstract}    
\input{sec/1_intro}
\input{sec/2_related_work}
\input{sec/3_method}
\input{sec/4_experiments}
\input{sec/5_discussion_and_conclusion}
{
    \small
    \bibliographystyle{ieeenat_fullname}
    \bibliography{main}
}

\input{supplementary/all}

\end{document}

%% file: sec/0_abstract.tex
\begin{abstract}
Deep neural networks trained with backpropagation have achieved outstanding performance in vision tasks but remain biologically implausible, computationally demanding, and difficult to interpret. The Forward-Forward (FF) algorithm offers a promising alternative by training each layer independently through local goodness objectives. However, its purely local optimization lacks hierarchical coordination across layers, and the decoupling of goodness from features leaves the representations unconstrained and semantically ambiguous. We propose a Hierarchical and Contrastive Learning FF framework (HCL-FF) to address these limitations. HCL-FF introduces (1) a coarse-to-fine hierarchical learning strategy that guides representations from low-level cues to high-level semantics, and (2) a supervised contrastive objective that enforces class-discriminative alignment after goodness decoupling. Experiments on CIFAR-10, CIFAR-100, and Tiny-ImageNet demonstrate that HCL-FF achieves new state-of-the-art performance among FF-based methods, with notable accuracy gains of $+5.46\%$, $+17.00\%$, and $+12.51\%$, respectively.
\end{abstract}

%% file: sec/1_intro.tex
\section{Introduction}
\label{sec:intro}

Backpropagation (BP)~\cite{rumelhart1986learning} has long been the foundation of modern deep learning.
Despite its empirical success, BP remains biologically implausible~\cite{grossberg1987competitive,crick1989recent, journe2022hebbian, lillicrap2020backpropagation}, as there is no evidence that the brain propagates error signals backward or stores neural activations for future updates.
It is also computationally demanding, requiring gradients and activations to be stored and propagated across all layers.
Moreover, it often yields opaque internal representations, limiting interpretability and transparency. 

These limitations have motivated growing interest in non-backpropagation learning paradigms~\cite{bartunov2018assessing, srinivasan2023forward, ernoult2022towards, ororbia2023backpropagation, journe2022hebbian, kohan2023signal}. 
Among them, the Forward-Forward (FF) algorithm~\cite{hinton2022forward} has emerged as a promising bottom-up alternative that trains each layer independently to maximize or minimize its \emph{goodness}, defined as the magnitude of the layer’s activations.
The layer-wise training scheme enables biologically plausible learning without gradient propagation across layers.
It also improves both parallelism~\cite{jaderberg2017decoupled, sun2025deeperforward} and memory efficiency~\cite{pau2023suitability}, as layers update without storing activations or waiting for global gradients, making it attractive for lightweight and edge settings~\cite{baghersalimi2023layer}. Finally, FF offers enhanced interpretability through direct visualization of layer-wise goodness responses~\cite{papachristodoulou2024convolutional}.

Despite these appealing properties, the FF algorithm faces two key limitations.
First, its greedy layer-wise learning lacks hierarchical coordination. While CNNs trained with backpropagation naturally learn a progression from low-level cues to high-level semantics, FF forces shallow layers to infer high-level semantic concepts directly, often leading to suboptimal early-layer representations.
Second, multi-layer learning remains ineffective due to the goodness decoupling problem. Specifically, FF decouples the goodness information from the layer output via vector-length normalization~\cite{hinton2022forward} or layer normalization~\cite{sun2025deeperforward}, which removes magnitude information and preserves only relative activation patterns, preventing subsequent layers from trivially inheriting the previous goodness signal.
However, the local goodness objective optimizes only the goodness score, i.e., the activation magnitude, leaving the goodness-decoupled features unconstrained and semantically ambiguous once the goodness is removed.

Recent studies attempt to mitigate the resulting information loss by relaxing strict decoupling, using batch normalization~\cite{papachristodoulou2024convolutional, dooms2023trifecta} or triangle activation function~\cite{chen2025self}. 
Although these modifications accelerate learning, they leak goodness information across layers, allowing deeper layers to exploit it rather than learn new patterns.
As a result, these methods tend to overfit and are limited to relatively shallow networks.
This reveals a fundamental \emph{decoupling dilemma} in FF learning:
Strict goodness decoupling prevents deeper layers from overfitting on previous goodness signals, yet it inevitably discards valuable semantic information encoded in activation magnitudes, the only part of the representation directly optimized by the goodness objective.

To address these limitations and resolve the decoupling dilemma, we propose a Hierarchical and Contrastive Learning FF framework (HCL-FF) that preserves the layer-wise nature of FF while introducing two key innovations.
First, we introduce a coarse-to-fine hierarchical learning strategy: shallow layers are supervised using coarse super-class targets, and deeper layers are progressively assigned finer-grained labels. This curriculum reduces early-layer complexity, improves gradient-free coordination across depth, and encourages representations to evolve from broad semantic structure to fine-grained discrimination.
Second, we incorporate supervised contrastive learning~\cite{khosla2020supervised} on the goodness-decoupled features. This explicitly constrains their relational geometry, aligning samples from the same class and separating different classes, thereby restoring the semantic structure that is lost when magnitude (goodness) is removed.
Extensive experiments show that HCL-FF sets a new state-of-the-art among FF-based methods, delivering higher accuracy, stronger feature separability, and stable semantics after goodness decoupling. Notably, HCL-FF achieves improvements of $+5.46\%$ on CIFAR-10, $+17.00\%$ on CIFAR-100, and $+12.51\%$ on Tiny-ImageNet, demonstrating the effectiveness of hierarchical coordination and contrastive grounding in overcoming long-standing challenges of FF learning.

%% file: sec/2_related_work.tex
\section{Related Work}
\label{sec:related_work}

\subsection{Learning Without Backpropagation}
Recent studies~\cite{ororbia2023brain} have explored alternatives to backpropagation, pursuing biologically plausible learning mechanisms. Feedback alignment and its variants~\cite{lillicrap2016random, nokland2016direct, akrout2019deep, frenkel2021learning, launay2019principled} replace the transposed forward weights used in backpropagation with random or learned feedback matrices to relax the weight-symmetry/transport constraint.
Target Propagation methods~\cite{bengio2014auto, lee2015difference, ernoult2022towards, bartunov2018assessing} approximate local targets through learned inverse mappings, allowing updates without backpropagated gradients. 
Hebbian-based methods~\cite{lagani2021hebbian, miconi2021hebbian, moraitis2022softhebb, journe2022hebbian} define weight updates from correlations between pre- and post-synaptic activity, often with competitive learning dynamics~\cite{rumelhart1985feature, grossberg1987competitive, song2000competitive}. Local representation alignment methods~\cite{ororbia2019biologically, ororbia2023backpropagation} compute the local alignment targets with error units and error weights. The forward-only framework PEPITA~\cite{dellaferrera2022error, srinivasan2023forward} replaces the backward pass with a second forward pass that perturbs inputs using error information, yielding Hebbian-like local updates.
Signal propagation~\cite{kohan2023signal} treats the target as part of the forward input, propagating both input and target jointly through the network to achieve single-pass learning.

\subsection{Forward-Forward Algorithm}
The Forward-Forward (FF) algorithm~\cite{hinton2022forward} replaces backpropagation with a layer-wise local goodness objective. Formally, given an input $x$, each layer computes
\begin{align}
&y^{(1)} = \phi(W^{(1)} x), \\
&y^{(\ell)} = \phi(W^{(\ell)} z^{{(\ell-1)}}), \quad \text{for} \quad 2 \leq \ell \leq L, \label{eq:ff_y} \\
&g^{(\ell)} = \sum_{i=1}^{N^{(\ell)}} (y_i^{(\ell)})^2, \quad 
z^{(\ell)} = \frac{y^{(\ell)}}{\sqrt{\frac{1}{N^{(\ell)}}g^{(\ell)} + \epsilon}}, \label{eq:ff_norm}
\end{align}
where $\phi$ denotes the activation function, $W^{(\ell)}$ is the weight matrix of layer $\ell$, $z^{{(\ell-1)}}$ is the feature from the previous layer, and $N^{(\ell)}$ is the number of neurons. The term $g^{(\ell)}$ represents the goodness, defined as the squared sum of activations. To prevent deeper layers from trivially inheriting the goodness signal, each layer normalizes its activations to produce a unit-length feature vector $z^{(\ell)}$, thereby decoupling goodness from the representation. During training, FF maximizes $g^{(\ell)}$ for positive samples (e.g., correct image–label pairs) and minimizes it for negative samples, encouraging layers to produce stronger activations for correct labels and suppress responses for incorrect ones. 

Since its introduction, the FF algorithm has inspired many extensions~\cite{giampaolo2023investigating, ororbia2023predictive, scodellaro2023training, reyes2024forward, lee2023symba, zhao2025cascaded, paliotta2023graph}.
However, existing variants largely remain limited by two core issues: (1) the lack of hierarchical coordination and (2) the decoupling dilemma.
CwComp~\cite{papachristodoulou2024convolutional} leverages the CIFAR-100 super-classes but uses them merely as a heuristic to handle larger label spaces.
To address coordination, Trifecta~\cite{dooms2023trifecta} introduces block-wise backpropagation, but this violates FF’s layer-wise independence.
Collaborative FF~\cite{lorberbom2024layer} adds a global goodness objective but sacrifices FF’s inherent parallelism, as layers must wait for a global signal before updating.
To address the decoupling issue, CwComp~\cite{papachristodoulou2024convolutional} and Trifecta~\cite{dooms2023trifecta} replace strict goodness decoupling with batch normalization, while SCFF~\cite{chen2025self} employs a triangle activation function~\cite{coates2011analysis}. These modifications ease optimization but leak goodness signal across layers, leading to overfitting and limiting scalability to deeper networks.
Recent contrastive FF variants~\cite{chen2025self, aghagolzadeh2024marginal, ahamed2023forward} apply contrastive objectives to raw activations rather than to goodness-decoupled features, leaving the semantic collapse after decoupling unresolved.
Most recently, DeeperForward~\cite{sun2025deeperforward} successfully trained a 17-layer CNN by redefining goodness as the mean of activations to mitigate neuron deactivation and applying layer normalization to strictly decouple goodness information, as shown in Eq.~\ref{eq:ff_mean}, where $\sigma$ denotes the standard deviation.
\begin{align}
&g^{(\ell)} = \frac{1}{N^{(\ell)}} \sum_{i=1}^{N^{(\ell)}} y_i^{(\ell)}, \quad 
z^{(\ell)} = \frac{y^{(\ell)} - g^{(\ell)}}{\sqrt{{(\sigma^{(\ell)})}^2 + \epsilon}}, \label{eq:ff_mean}
\end{align}
While this restores strict decoupling and improves depth scalability, it still suffers from the loss of semantic meaning in the representations after goodness decoupling.

%% file: sec/3_method.tex
\begin{figure*}[t]
  \centering
  \includegraphics[width=0.985\textwidth]{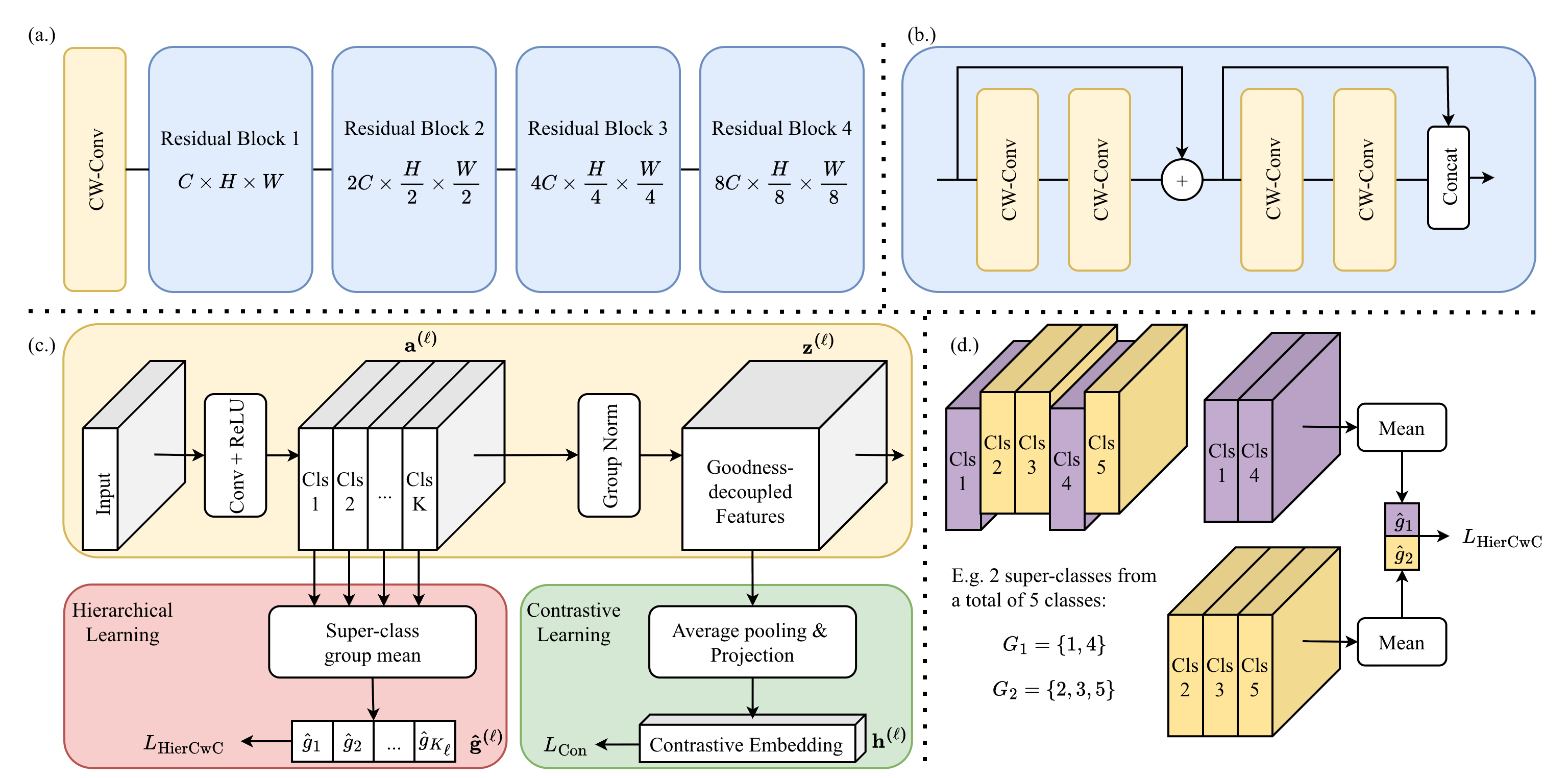}
  \caption{
    (a) Overall architecture.
    (b) Structure of a residual block.
    (c) CW-Conv layer. Channels are partitioned into $K$ class-specific subsets, which are further grouped into super-classes for computing super-class mean-goodness. Subset-wise normalization decouples the goodness signal, and a supervised contrastive loss on the goodness-decoupled features provides semantic grounding.
    (d) Illustration of hierarchical learning. Super-class goodness $\hat{g}$ is obtained by averaging the per-class goodness within each super-class group.
  }
  \label{fig:overview}
\end{figure*}

\section{Methodology}
\label{sec:method}
In this section, we first describe the baseline Forward-Forward architecture inherited from DeeperForward~\cite{sun2025deeperforward}, which serves as the foundation of our method. 
We then introduce our two key contributions: Hierarchical Learning and Contrastive Learning. 
Finally, we present the overall HCL-FF framework that integrates both objectives.

\subsection{Baseline Forward-Forward Architecture}
\label{sec:FF}
We adopt the FF backbone from DeeperForward~\cite{sun2025deeperforward} and briefly summarize it here for completeness.
The architecture is built upon Channel-Wise Convolution (CW-Conv) layers. As illustrated in Fig.~\ref{fig:overview}(a), the network begins with a CW-Conv stem and proceeds through four residual blocks, each containing four CW-Conv layers connected by residual shortcuts to enhance representational richness.

Each CW-Conv layer operates as a standard convolution followed by ReLU activation and is trained independently without gradient propagation across layers. 
Following the Channel-wise Competitive (CwC) formulation~\cite{papachristodoulou2024convolutional}, the output activations are partitioned into $K$ subsets along the channel dimension, where $K$ denotes the number of classes. 
Formally, an activation tensor $a^{(\ell)} \in \mathbb{R}^{C \times H \times W}$ at layer $\ell$ is reshaped into ${a^{(\ell)}}' \in \mathbb{R}^{K \times C' \times H \times W}$ where $C' = C/K$. 
In practice, we ensure that $C$ is divisible by $K$.
Each subset ${a^{(\ell)}}'_{k} \in \mathbb{R}^{C' \times H \times W}$ represents the activation responses associated with class $k$. Following the mean-goodness formulation in Eq.~\ref{eq:ff_mean}, the per-class mean-goodness is computed by averaging activations within each subset, yielding goodness $g^{(\ell)} \in \mathbb{R}^{K}$. 
The CwC loss is then defined as the softmax cross-entropy over the per-class mean-goodness values, which maximizes the goodness of the ground-truth class while suppressing others:
\begin{align}
& L^{(\ell)}_{\text{CwC}} (g^{(\ell)}, y) = -\sum_{i=1}^{K} y_i \log \frac{\exp(g^{(\ell)}_i)}{\sum_{j=1}^{K} \exp(g^{(\ell)}_j)}
\label{eq:cwc_loss}
\end{align}

To obtain the goodness-decoupled feature $z^{(\ell)}$ following Eq.~\ref{eq:ff_mean}, GroupNorm is applied to activation tensor $a^{(\ell)}$ with the number of groups set to $K$, which is equivalent to normalizing independently within each subset ${a^{(\ell)}}'_{k} \in \mathbb{R}^{C' \times H \times W}$. This per-group normalization removes the global magnitude component used as the goodness signal while preserving relative activation patterns within each class group.
The goodness-decoupled feature $z^{(\ell)}$ is then forwarded to the next layer, ensuring that subsequent layers cannot trivially exploit the goodness information from previous layers. 
At inference time, the Signal Integrating and Pruning Module~\cite{sun2025deeperforward} is applied to select the optimal layer interval $[s,e]$ on the validation set. The goodness scores within this interval are averaged as $\tilde g = \frac{1}{e-s+1}\sum_{\ell=s}^{e} g^{(\ell)}$, and the final class prediction is obtained by taking the $\arg\max$ over the ensembled goodness scores $\tilde g$.

\subsection{Hierarchical Learning}
\label{sec:hierarchical}

\begin{figure}[t]
    \centering
    \includegraphics[width=\columnwidth]{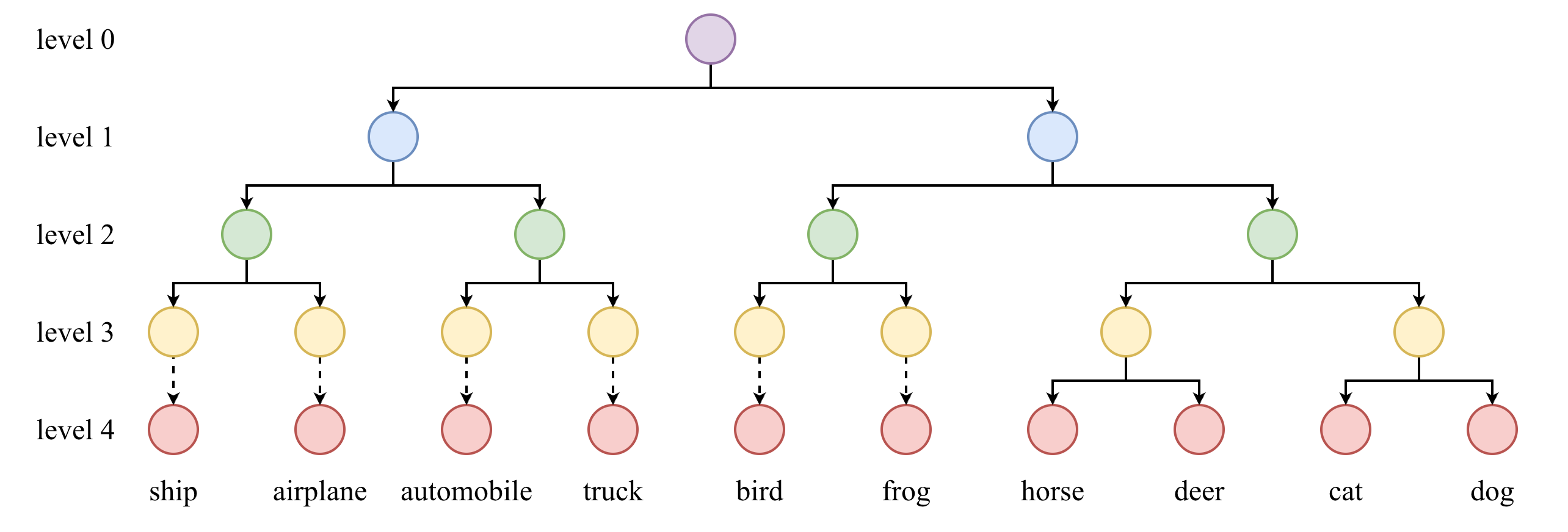}
    \caption{
    Hierarchy of CIFAR-10 classes constructed via clustering on class prototypes derived from a pre-trained classifier.
    Each level represents a different granularity of super-class grouping.
    }
    \label{fig:hierarchy}
\end{figure}

The design above completes the layer-wise FF learning pipeline. However, this purely local optimization lacks coordination across layers.
Moreover, shallow layers are forced to discriminate among all $K$ classes directly, imposing an overly complex objective that destabilizes early learning and yields suboptimal representations.

To address this limitation, we introduce a Hierarchical Learning paradigm that organizes supervision in a coarse-to-fine manner.
Shallow layers focus on distinguishing broad semantic groups, while deeper layers progressively refine their representations toward fine-grained distinctions. 
Formally, let $\mathcal{Y} = \{1, \dots, K\}$ denote the set of fine-grained classes. For layer $\ell$, we define a set of super-classes as $\{G^{(\ell)}_1, \dots, G^{(\ell)}_{K_\ell}\}$, where $K_\ell$ denotes the number of super-classes at level $\ell$ and $\{G^{(\ell)}_1, \dots, G^{(\ell)}_{K_\ell}\}$ forms a partition of $\mathcal{Y}$.
We set $K_1 \leq K_2 \leq \dots \leq K_L = K$ to establish a coarse-to-fine curriculum. Given the per-class goodness responses $g^{(\ell)} = (g^{(\ell)}_1, \dots, g^{(\ell)}_K)$, the super-class-level goodness is obtained by averaging the goodness within each super-class, as illustrated in Fig.~\ref{fig:overview}(d):
\begin{align}
\hat{g}^{(\ell)}_j = \frac{1}{|G^{(\ell)}_j|} \sum_{i \in G^{(\ell)}_j} g^{(\ell)}_i, \quad j = 1,2, \dots, K_\ell
\end{align}
By substituting the fine-grained labels and goodness in Eq.~\ref{eq:cwc_loss} with the super-class counterparts $\hat{y}^{(\ell)}$ and $\hat{g}^{(\ell)}$, we derive the Hierarchical Channel-Wise Competitive (HierCwC) loss, which enables hierarchical learning while preserving the layer-wise independence of FF training:
\begin{align}
&L^{(\ell)}_{\text{HierCwC}} (\hat{g}^{(\ell)}, \hat{y}^{(\ell)}) = -\sum_{i=1}^{K_{\ell}} \hat{y}^{(\ell)}_i \log \frac{\exp(\hat{g}^{(\ell)}_i)}{\sum_{j=1}^{K_\ell} \exp(\hat{g}^{(\ell)}_j)} \label{eq:hier_loss}
\end{align}

To compute the super-class partitions ${G^{(\ell)}_j}$, we require a class hierarchy that defines semantic relationships among fine-grained labels.
The class hierarchy can be constructed in various ways, such as using external taxonomies~\cite{miller1995wordnet, deng2009imagenet}, semantic word embeddings~\cite{mikolov2013efficient}, or data-driven clustering~\cite{wan2020nbdt}. 
In this work, we adopt the data-driven approach~\cite{wan2020nbdt}.
Specifically, we first pre-train the model without hierarchical supervision using the CwC objective in Eq.~\ref{eq:cwc_loss}. Then, we freeze the model and train a linear classifier on the global-average-pooled feature of the final layer.
Each row of the trained classifier weight matrix serves as a class prototype in the learned feature space.
We then $\ell_2$-normalize these prototypes and perform hierarchical agglomerative clustering to construct a tree that captures semantic similarity among classes.

An example of the resulting hierarchy for CIFAR-10 is shown in Fig.~\ref{fig:hierarchy}, where each internal node represents a super-class.
Once constructed, network layers are mapped to selected tree depths to define the super-class groupings.
The mapping between layers and tree depths defines the learning curriculum, following two principles: (1) the supervision depths increase monotonically with network depth, ensuring a coarse-to-fine learning process, and (2) the final layer corresponds to the leaf level, enabling fine-grained classification.
To maintain a valid partition of the class set $\mathcal{Y}$ at every level, each node that terminates before the maximum tree depth is extended by duplicating itself as its own descendant, illustrated by the dashed lines in Fig.~\ref{fig:hierarchy}, thereby preserving coverage and non-overlap of $\mathcal{Y}$.

\subsection{Contrastive Learning}
\label{sec:contrastive}
A second major limitation of FF algorithms is the \emph{decoupling dilemma}.
On one hand, without goodness decoupling, deeper layers can trivially inherit the goodness signal from preceding layers, preventing new feature learning and leading to overfitting.
On the other hand, once goodness is removed, the learned features lose their semantic meaning.
Because the standard FF optimizes only the goodness score, the training objective constrains the activation magnitude but leaves relative activation patterns unconstrained. 
As a result, goodness-decoupled features become semantically ambiguous, undermining interpretability and weakening learning across layers.

To resolve this dilemma, we introduce a contrastive objective applied directly to the goodness-decoupled features, explicitly constraining their representation and preserving semantic meaning.
Specifically, at each CW-Conv layer, the goodness-decoupled feature $z^{(\ell)}$ is global-average-pooled and projected to a latent embedding $\mathbf{h}^{(\ell)}$ through a linear projection head. 
We then apply a supervised contrastive loss~\cite{khosla2020supervised} that encourages embeddings from the same class to cluster together while pushing those from different classes apart. 
Formally, given a mini-batch of $N$ samples with embeddings ${\{\mathbf{h}^{(\ell)}_i\}}_{i=1}^{N}$ and labels ${\{y_i\}}_{i=1}^{N}$, the contrastive loss at layer $\ell$ is defined as
\begin{equation}
\begin{aligned}
L_{\text{Con}}^{(\ell)}
&= \sum_{i=1}^{N}
\frac{-1}{|\mathcal{P}(i)|}
\sum_{p\in\mathcal{P}(i)}
\log \frac{
\exp(\mathrm{sim}(\mathbf{h}^{(\ell)}_i,\mathbf{h}^{(\ell)}_p)/\tau)
}{
\sum_{a\in A(i)} \exp(\mathrm{sim}(\mathbf{h}^{(\ell)}_i,\mathbf{h}^{(\ell)}_a)/\tau)
}
\end{aligned}
\label{eq:con_loss}
\end{equation}
where $A(i) = \{1, \dots, N\} \setminus \{i\}$ denotes all indices except $i$, $\mathcal{P}(i) = \{a \in A(i): y_a = y_i\}$ represents the set of samples sharing the same class as sample $i$, $\text{sim}(\cdot, \cdot)$ denotes cosine similarity, and $\tau$ is a temperature parameter.

We emphasize that the proposed contrastive objective is applied directly to the goodness-decoupled representation $z^{(\ell)}$, rather than to the raw activations or the goodness scores. By operating explicitly in this goodness-decoupled feature space, the contrastive objective preserves the semantic meaning of $z^{(\ell)}$, thereby addressing the decoupling dilemma.
Intuitively, goodness governs the scale of activations, while contrastive learning constrains their relative direction, ensuring that goodness-decoupled features remain semantically meaningful and discriminative.

\subsection{HCL-FF Framework}
The overall objective integrates the hierarchical objective from Eq.~\ref{eq:hier_loss} and the contrastive objective from Eq.~\ref{eq:con_loss}:
\begin{align}
L_{\text{total}}^{(\ell)} =
L_{\text{HierCwC}}^{(\ell)} + \lambda L_{\text{Con}}^{(\ell)},
\label{eq:total_loss}
\end{align}
where $\lambda$ is a weighting coefficient set to $1$ in all experiments.
The two objectives play complementary roles: the hierarchical loss provides structured coarse-to-fine supervision across depth, while the contrastive loss preserves the semantic geometry of the goodness-decoupled features.  
Importantly, contrastive learning always uses fine-grained labels, as replacing them with super-class labels would collapse meaningful intra-group distinctions.

To realize hierarchical supervision in practice, we map the 17 layers in our architecture to hierarchy levels according to the following rule:
\begin{align}
\text{level}(0) = 1, \quad 
\text{level}(i) = \left\lceil \dfrac{i \, (D-1)}{16} \right\rceil, \quad i = 1, \dots, 16,
\end{align}
where $D$ is the maximum tree depth. This mapping ensures that early layers receive coarse supervision, while deeper layers progressively transition to fine-grained targets.

The hierarchical curriculum also enhances the effectiveness of data augmentation. 
In prior FF methods, shallow layers faced overly complex fine-grained discrimination, diminishing the benefits of augmentation.
By assigning coarse objectives to early layers, our hierarchical formulation reduces this burden, allowing augmentation to meaningfully enrich feature diversity and improve robustness.
This enables the use of strong data augmentation in practice, including random cropping, horizontal flipping, color jittering, and random grayscale conversion.

Together, these components form an effective FF framework that addresses two core limitations of existing FF methods: the lack of hierarchical coordination and the decoupling dilemma, leading to the consistent performance improvements demonstrated in Sec.~\ref{sec:experiments}.

%% file: sec/4_experiments.tex
\section{Experiments}
\label{sec:experiments}

\input{tables/benchmark}

\subsection{Experimental Setting}

We conduct experiments on five standard benchmarks: CIFAR-10, CIFAR-100~\cite{krizhevsky2009learning}, MNIST~\cite{lecun2002gradient}, Fashion-MNIST~\cite{xiao2017fashion}, and Tiny-ImageNet~\cite{Le2015TinyIV}. 
Tiny-ImageNet (64$\times$64, 200 classes) provides a higher-resolution and more challenging setting, while the remaining datasets cover diverse visual domains and class granularities. 
Models are trained for 1000 epochs on Tiny-ImageNet, CIFAR-100, and CIFAR-10, and for 150 epochs on Fashion-MNIST and MNIST. 
We use a batch size of 512 for Tiny-ImageNet and CIFAR-100, and 128 for the remaining datasets. 
We use the Adam optimizer with weight decay $1 \times 10^{-4}$, and a cosine-annealing learning rate schedule decaying from $8\times10^{-2}$ to $2\times10^{-4}$. 
All experiments are conducted on a single NVIDIA RTX~A6000 GPU. 
Detailed network configurations, optimization hyperparameters, and data augmentation parameters are provided in the Appendix.

\subsection{General Benchmarks}
\input{tables/tinyimagenet200}

We compare against representative non-backpropagation (non-BP) methods~\cite{dellaferrera2022error, ernoult2022towards, ororbia2023backpropagation, journe2022hebbian, kohan2023signal}, standard backpropagation (BP) approaches~\cite{he2016deep}, and Forward-Forward (FF) baselines~\cite{hinton2022forward, lee2023symba, zhao2025cascaded, papachristodoulou2024convolutional, dooms2023trifecta, chen2025self, sun2025deeperforward}.
To ensure a fair comparison with BP-based models, we implement a wider ResNet-20 whose parameter count matches that of our FF architecture. Table~\ref{tab:benchmark} reports results on CIFAR-10, CIFAR-100, MNIST, and Fashion-MNIST.
HCL-FF achieves state-of-the-art performance among all FF-based models.
Compared to DeeperForward~\cite{sun2025deeperforward}, our method provides substantial gains of $+5.46\%$ on CIFAR-10 and $+17.00\%$ on CIFAR-100, highlighting the effectiveness of hierarchical coordination and contrastive grounding, particularly on datasets with larger class spaces.
Notably, HCL-FF reaches $91.68\%$ on CIFAR-10 and $70.09\%$ on CIFAR-100, surpassing the standard BP-trained ResNet-20~\cite{he2016deep} and narrowing the gap to the matched-parameter BP variant.
On simpler datasets such as MNIST and Fashion-MNIST, HCL-FF achieves $99.65\%$ and $93.87\%$, confirming strong generalization across visual complexity levels.
In comparison to other non-BP approaches, HCL-FF consistently achieves superior or comparable accuracy, reinforcing the viability of FF-based learning as an effective and biologically plausible training paradigm.

To further assess scalability, we extend our experiments to the more challenging Tiny-ImageNet dataset.
As shown in Table~\ref{tab:tiny}, HCL-FF significantly outperforms prior FF methods by $12.51\%$, demonstrating improved capacity to handle higher-resolution images and larger label spaces under strict layer-wise training.
Overall, these results show that hierarchical supervision and contrastive grounding jointly address long-standing limitations of FF learning, delivering substantial performance gains while preserving fully layer-local training.

\subsection{Ablation Studies}
\input{tables/ablation}

To evaluate the contribution of each component in HCL-FF, we evaluate six model variants on CIFAR-10 and CIFAR-100; results are summarized in Table~\ref{tab:ablation}.

Introducing the contrastive objective alone (V2$\rightarrow$V3) yields notable improvements of $+2.25\%$ on CIFAR-10 and $+9.12\%$ on CIFAR-100. This confirms that contrastive supervision stabilizes goodness-decoupled features and prevents semantic drift by explicitly constraining their relational geometry. The much larger gain on CIFAR-100 reflects its higher class granularity, where unconstrained features collapse more severely after goodness removal.

Adding the hierarchical objective alone (V2$\rightarrow$V4) provides a substantial gain on CIFAR-100 ($+6.96\%$), while slightly reducing CIFAR-10 accuracy ($-0.60\%$). This suggests that the benefit of coarse-to-fine supervision is more pronounced in larger label spaces.

Data augmentation alone brings only modest gains (V1$\rightarrow$V2), indicating that without hierarchical structure, shallow layers remain overburdened by fine-grained discrimination and cannot fully leverage augmented variability.
However, when hierarchical supervision is present (V5$\rightarrow$V6), augmentation yields significantly larger gains ($+3.39\%$ and $+5.54\%$).
This shows that simplifying early objectives enables augmented samples to meaningfully enrich feature diversity and improve robustness.

Finally, combining all components (V6) achieves the strongest overall performance. These results highlight the complementary nature of hierarchical coordination, contrastive grounding, and strong data augmentation in overcoming the limitations of layer-wise FF training.

\begin{figure}[t]
  \centering
  \includegraphics[width=\columnwidth]{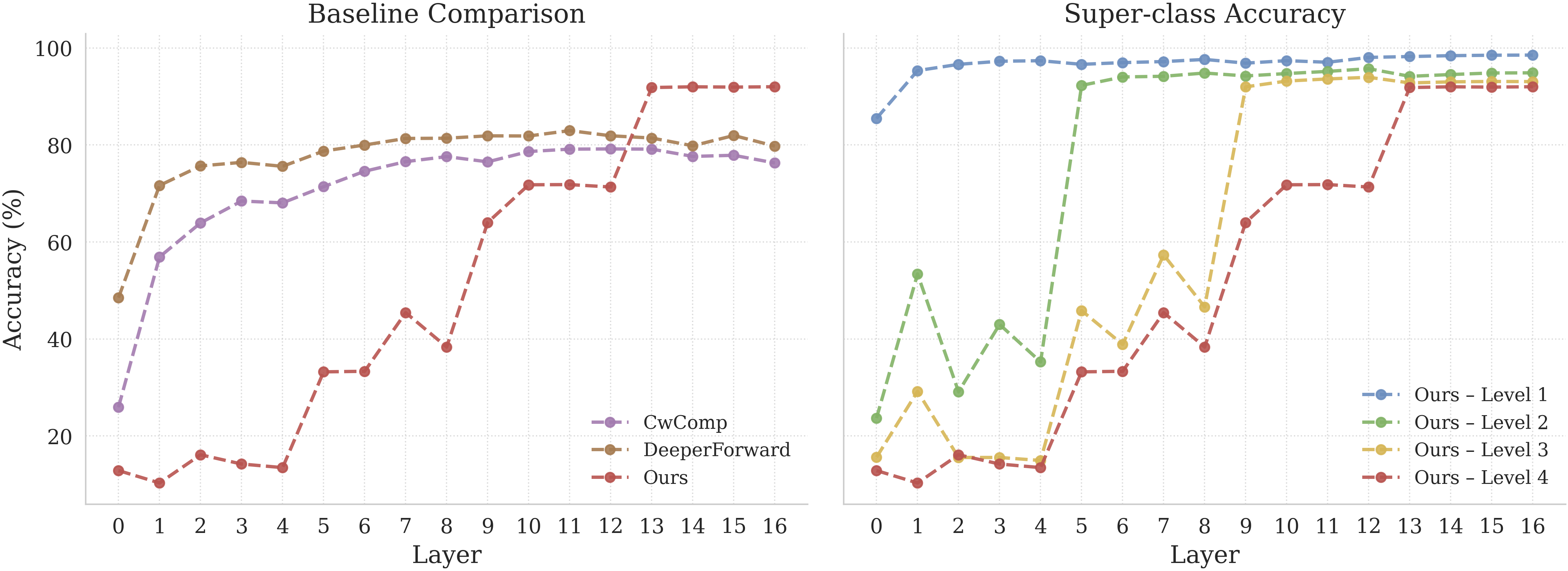}
  \caption{
    Layer-wise classification accuracy on CIFAR-10. 
    The left panel compares our HCL-FF with prior FF-based models using per-class goodness at each layer.
    The right panel reports the super-class accuracy of HCL-FF at each hierarchy level defined in Fig.~\ref{fig:hierarchy}.
    Predictions are obtained by taking the $\arg\max$ over the class-wise or super-class-wise goodness responses at each layer.
  }
  \label{fig:layer_acc}
\end{figure}

\begin{figure}[t]
    \centering
    \includegraphics[width=\columnwidth]{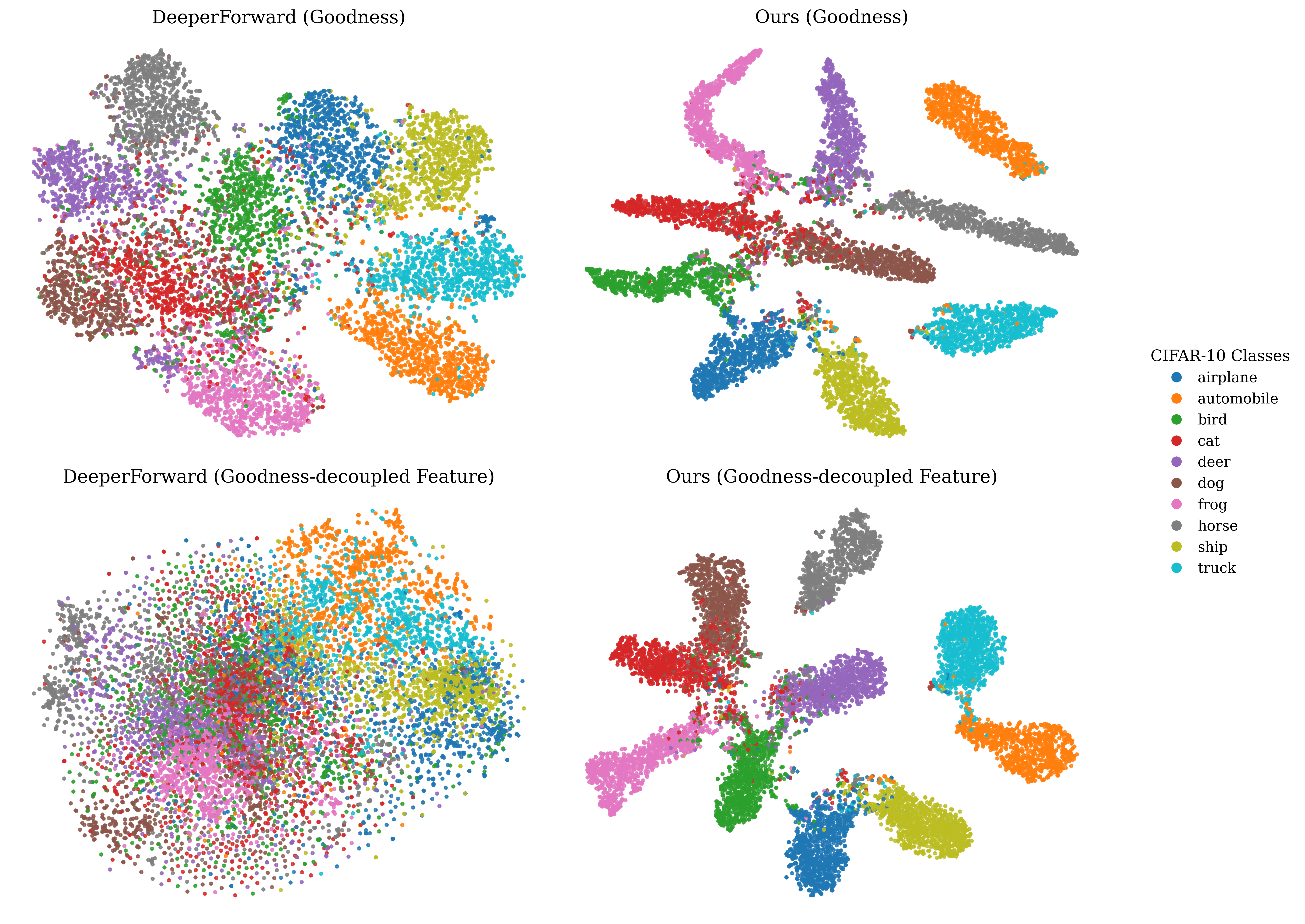}
    \caption{
    t-SNE visualization of goodness and goodness-decoupled features on CIFAR-10. 
    }
    \label{fig:tsne}
\end{figure}

\subsection{Effect of Different Hierarchies}
\input{tables/hierarchy}
To examine the sensitivity of HCL-FF to the choice of hierarchical structures, we compare three ways of constructing the class hierarchies:
(1) WordNet-based hierarchy derived from human semantic taxonomies~\cite{miller1995wordnet},
(2) Word2Vec-based hierarchy obtained by hierarchical agglomerative clustering of class-name embeddings~\cite{mikolov2013efficient}, and
(3) Data-driven hierarchy induced from classifier weights as described in Sec.~\ref{sec:hierarchical}.

As shown in Table~\ref{tab:hierarchy}, the WordNet hierarchy achieves the highest accuracy, likely because its groupings closely reflect human-defined visual semantics.
The data-driven hierarchy performs comparably without relying on external priors, demonstrating strong adaptability for datasets lacking predefined taxonomies. 
The Word2Vec-based clustering slightly underperforms, suggesting that linguistic similarity does not always align with visual similarity in fine-grained recognition. 
Together, these results show that HCL-FF is robust to the hierarchy construction method and can effectively leverage both semantic and data-driven structures, making it broadly applicable across datasets with or without existing taxonomies.

\begin{figure*}[t]
    \centering
    \includegraphics[width=0.95\textwidth]{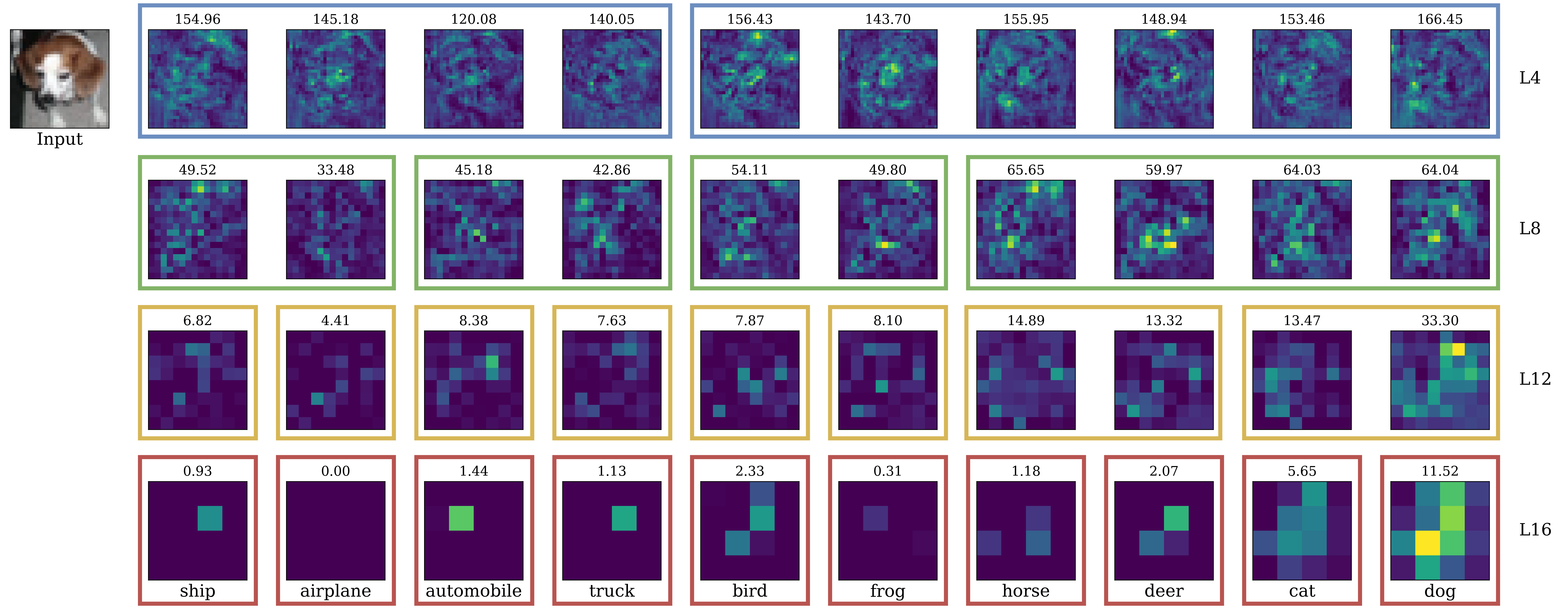}
    \caption{
    Visualization of class-wise goodness responses from the final layer of each residual block on the CIFAR-10 dataset. Each column corresponds to the mean goodness map of the channel subset associated with a given class, with the mean goodness value displayed above each map. Color frames correspond to the super-class grouping in Fig.~\ref{fig:hierarchy}.
    As depth increases, responses become semantically aligned with the target class. The classes within the same super-class generally have higher goodness responses, showing a coarse-to-fine progression.
    }
    \label{fig:heatmap}
\end{figure*}

\subsection{Analysis of Decoupling Dilemma}

The FF algorithm inherently faces a decoupling dilemma: while removing the goodness signal is essential to prevent goodness leakage across layers, it also risks removing semantic meaning from the decoupled features.
We investigate this dilemma through three complementary analyses.

\paragraph{Per-layer classification accuracy.}
Fig.~\ref{fig:layer_acc} reports layer-wise classification accuracy. 
The left panel computes predictions via the $\arg\max$ over per-class goodness at each layer, while the right panel reports super-class accuracy following the hierarchy in Fig.~\ref{fig:hierarchy}.
CwComp saturates after layer~8, indicating that goodness signal leaks across layers and allows deeper layers to overfit rather than learn new discriminative structure.
DeeperForward alleviates this issue through strict goodness decoupling, yet its accuracy still plateaus early. This suggests that while leakage is prevented, the goodness-decoupled features become unconstrained and drift semantically, offering limited discriminative improvement with depth.
In contrast, HCL-FF maintains steady accuracy gains across layers, demonstrating that hierarchical and contrastive supervision jointly enable meaningful feature refinement throughout the network.
The right panel further reveals a clear coarse-to-fine pattern: shallow layers reliably capture broad super-class distinctions, while deeper layers progressively specialize toward fine-grained categories.

\paragraph{Linear-probe evaluation.}

\input{tables/linear_probe.tex}
To quantitatively assess feature semantics, we freeze the model and train a linear classifier on the global-average-pooled final-layer features, both before and after normalization.
As shown in Table~\ref{tab:linear_probe}, DeeperForward, which strictly removes goodness following Eq.~\ref{eq:ff_mean}, suffers a pronounced accuracy drop after normalization, indicating that a large portion of its discriminative power is carried by the magnitude component. Once removed, the remaining feature collapses.
CwComp shows a minor change after normalization, but its absolute accuracy is lower than DeeperForward. This reflects its weaker decoupling: batch normalization leaks goodness across layers, allowing the model to retain magnitude information. However, this leakage ultimately harms learning by encouraging overfitting rather than meaningful feature development.
In contrast, HCL-FF maintains nearly identical accuracy before and after normalization while achieving the highest performance. This shows that HCL-FF both enforces strict goodness decoupling and preserves the semantic structure of the goodness-decoupled features via contrastive regularization, effectively resolving the decoupling dilemma.

\paragraph{t-SNE visualization.}
Figure~\ref{fig:tsne} visualizes the learned embeddings on CIFAR-10 using t-SNE~\cite{maaten2008visualizing}, with points colored by class.
We plot both the ten-dimensional goodness response and the corresponding goodness-decoupled representations from the final layer, i.e., $g^{(16)}$ and $z^{(16)}$. 
In DeeperForward, the goodness representations are well-separated across classes, indicating that the model effectively learns to encode discriminative magnitude information.
However, after normalization, the goodness-decoupled features collapse and lose inter-class structure, suggesting that once the magnitude is removed, the remaining relative patterns are unconstrained and prone to semantic drift.
In contrast, HCL-FF preserves clear class separation in both $g^{(16)}$ and $z^{(16)}$, indicating that the combined hierarchical and contrastive objectives constrain both the goodness magnitudes and the geometry of goodness-decoupled features.

\subsection{Qualitative Results}

Unlike standard deep networks that typically require post-hoc interpretability tools, the FF framework naturally exposes class-wise goodness responses at every layer, providing built-in transparency into how evidence evolves with depth.
Figure~\ref{fig:heatmap} visualizes the class-wise goodness responses from the final layer of each residual block for an example image from CIFAR-10.
Each column corresponds to the mean goodness map of the channel subset associated with a given class, while frame colors indicate the super-class groupings defined by the hierarchy in Fig.~\ref{fig:hierarchy}. 
The mean goodness value is displayed above each map.

At shallow layers (e.g., L4 and L8), the responses are spatially diffuse and semantically coarse, activating for multiple classes that share similar low-level features such as colors and edges.
As depth increases (L12 and L16), the responses become more localized and strongly biased toward the target class, reflecting the emergence of fine-grained semantic selectivity.
Within each layer, classes belonging to the same super-class exhibit higher relative goodness, consistent with the hierarchical grouping and the coarse-to-fine curriculum used during training.
These qualitative trends visually corroborate our quantitative findings, confirming that HCL-FF exhibits a clear coarse-to-fine progression of semantic refinement across layers.

%% file: tables/benchmark.tex
\begin{table*}[t]
\centering
\caption{Classification performance on CIFAR-10, CIFAR-100, MNIST, and Fashion-MNIST (F-MNIST). Mean and standard deviation of five runs are reported. The best accuracy for each type is highlighted in bold. $^\dagger$: Reproduced results.}
\resizebox{0.8\textwidth}{!}{
\begin{tabular}{l l c c c c c}
\toprule
\textbf{Type} & \textbf{Method} & \textbf{Arch.} & \textbf{CIFAR-10} & \textbf{CIFAR-100} & \textbf{MNIST} & \textbf{F-MNIST} \\
\midrule
\multirow{5}{*}{non-BP} 
 & PEPITA \cite{dellaferrera2022error} & CNN & $56.33 \pm 1.35$ & $27.56 \pm 0.60$ & $98.29 \pm 0.13$ & - \\
 & DTP \cite{ernoult2022towards} & CNN & $89.38 \pm 0.20$ & - & $98.93 \pm 0.04$ & $\mathbf{90.91 \pm 0.17}$ \\
 & rec-LRA \cite{ororbia2023backpropagation} & CNN & $\mathbf{93.88}$ & - & $98.18$ & $88.13$ \\
 & SoftHebb \cite{journe2022hebbian} & SoftHebb & $80.31 \pm 0.14$ & $56.0$ & $\mathbf{99.35 \pm 0.03}$ & - \\
 & SigProp \cite{kohan2023signal} & CNN & $91.66$ & $\mathbf{65.7}$ & - & - \\
\midrule
\multirow{2}{*}{BP} 
 & ResNet20$^\dagger$ \cite{he2016deep} & CNN & $91.25$ & $67.20$ & $\mathbf{99.64}$ & $93.34$ \\
 & ResNet20-Wide$^\dagger$ \cite{he2016deep} & CNN & $\mathbf{93.98}$ & $\mathbf{76.72}$ & $99.63$ & $\mathbf{93.89}$ \\
\midrule
\multirow{8}{*}{FF} 
 & FF \cite{hinton2022forward} & MLP & $59.00$ & $18.15$ & $98.69$ & - \\
 & SymBa \cite{lee2023symba} & MLP & $59.09$ & $29.28$ & $98.58$ & - \\
 & CaFo \cite{zhao2025cascaded} & CNN & $67.43$ & $40.76$ & $98.80$ & - \\
 & CwComp \cite{papachristodoulou2024convolutional} & CNN & $78.11 \pm 0.44$ & $51.23$ & $99.42 \pm 0.08$ & $92.31 \pm 0.32$ \\
 & Trifecta \cite{dooms2023trifecta} & CNN & $83.51 \pm 0.78$ & $35.26 \pm 0.23$ & $99.58 \pm 0.06$ & $91.44 \pm 0.49$ \\
 & SCFF \cite{chen2025self} & CNN & $80.75 \pm 0.12$ & - & $99.37 \pm 0.06$ & - \\
 & DeeperForward \cite{sun2025deeperforward} & CNN & $86.22 \pm 0.17$ & $53.09 \pm 0.79$ & $99.63 \pm 0.04$ & $93.13 \pm 0.13$ \\
 & Ours & CNN & $\mathbf{91.68 \pm 0.19}$ & $\mathbf{70.09 \pm 0.15}$ & $\mathbf{99.65 \pm 0.04}$ & $\mathbf{93.87 \pm 0.24}$ \\
\bottomrule
\end{tabular}}
\label{tab:benchmark}
\end{table*}

%% file: tables/tinyimagenet200.tex
\begin{table}[t]
\centering
\small
\setlength{\tabcolsep}{4pt}
\caption{Top-1 accuracy (\%) on Tiny-ImageNet. \\
$^\dagger$: Reproduced results.}
\resizebox{0.9\columnwidth}{!}{
\begin{tabular}{lcccc}
\toprule
\textbf{Method} & \textbf{ResNet-BP$^\dagger$} & \textbf{SCFF} & \textbf{DeeperForward$^\dagger$} & \textbf{Ours} \\
\midrule
Accuracy & 64.40 & 35.67 & 35.95 & 48.46 \\
\bottomrule
\end{tabular}
}
\label{tab:tiny}
\end{table}

%% file: tables/ablation.tex
\begin{table}[t]
\centering
\caption{Ablation study. "Hier." denotes hierarchical learning, "Con." denotes contrastive learning, 
and "Aug." denotes strong data augmentation. 
}
\resizebox{0.9\columnwidth}{!}{
\begin{tabular}{c c c c c c}
\toprule
\textbf{} &
\textbf{Hier.} & 
\textbf{Con.} & 
\textbf{Aug.} & 
\textbf{CIFAR-10 Acc.} &
\textbf{CIFAR-100 Acc.} \\
\midrule
V1 & $\times$ & $\times$ & $\times$ & 86.22 & 53.09 \\
V2 & $\times$ & $\times$ & \checkmark & 88.80 & 54.65 \\
V3 & $\times$ & \checkmark & \checkmark & 91.05 & 63.77 \\
V4 & \checkmark & $\times$ & \checkmark & 88.20 & 61.61 \\
V5 & \checkmark & \checkmark & $\times$ & 88.44 & 65.22 \\
V6 & \checkmark & \checkmark & \checkmark & \textbf{91.83} & \textbf{70.76} \\
\bottomrule
\end{tabular}
}
\label{tab:ablation}
\end{table}

%% file: tables/hierarchy.tex
\begin{table}[t]
\centering
\small
\setlength{\tabcolsep}{4pt}
\caption{Comparison of different hierarchy construction methods.}
\resizebox{0.8\columnwidth}{!}{
\begin{tabular}{lccc}
\toprule
\textbf{Method} & \textbf{WordNet} & \textbf{Word2Vec} & \textbf{Data-driven} \\
\midrule
CIFAR-100 Acc. & 71.01 & 69.59 & 70.76 \\
\bottomrule
\end{tabular}
}
\label{tab:hierarchy}
\end{table}

%% file: tables/linear_probe.tex
\begin{table}[t]
\centering
\small
\setlength{\tabcolsep}{4pt}
\caption{Linear-probe evaluation on the last layer's features before/after normalization.}
\resizebox{\columnwidth}{!}{
\begin{tabular}{lccc}
\toprule
\textbf{Method} & \textbf{CwComp} & \textbf{DeeperForward} & \textbf{Ours} \\
\midrule
CIFAR-10 Acc. & 76.29 / 76.64 & 84.46 / 76.93 & 91.93 / 91.91 \\
CIFAR-100 Acc. & 41.98 / 36.41 & 48.38 / 35.47 & 67.42 / 65.85 \\
\bottomrule
\end{tabular}
}
\label{tab:linear_probe}
\end{table}

%% file: sec/5_discussion_and_conclusion.tex
\section{Discussion and Conclusion}
\label{sec:discussion_and_conclusion}

\paragraph{Limitations.}
While HCL-FF advances FF learning, it still inherits certain constraints.
In our experiments, layer-wise optimization still converges more slowly and generalizes less effectively than end-to-end backpropagation.
Moreover, the CwC loss requires allocating channel subsets per class, causing channel count to scale with the number of classes.
Our HierCwC loss alleviates this by allowing shared channels in shallow layers, yet scaling to datasets with extremely large label spaces remains challenging.

\paragraph{Conclusion.}
We presented HCL-FF, a Hierarchical and Contrastive Learning FF framework that addresses two central challenges of FF learning: the lack of hierarchical coordination and the decoupling dilemma.
By integrating hierarchical supervision with contrastive grounding, HCL-FF learns semantically structured and progressively refined representations while maintaining fully layer-wise independence.
Our experiments across five benchmarks show state-of-the-art FF performance. These results highlight the potential of HCL-FF as a step toward more scalable, effective, and biologically plausible learning paradigms.

%% file: supplementary/all.tex
\section{Implementation Details}
\label{supp::arch}

\begin{table*}[t]
\centering
\caption{Architectural configurations and training hyperparameters for all datasets.}
\label{tab:arch_hparams}
\resizebox{\textwidth}{!}{
\begin{tabular}{lccccc}
\toprule
 & \textbf{CIFAR-10} & \textbf{CIFAR-100} & \textbf{MNIST} & \textbf{F-MNIST} & \textbf{Tiny-ImageNet} \\
\midrule
Channels per residual block & [100, 200, 400, 800] & [100, 200, 400, 800] & [40, 80, 160, 320] & [40, 80, 160, 320] & [200, 400, 800, 1600] \\
Batch size & 128 & 512 & 128 & 128 & 512 \\
Optimizer & Adam & Adam & Adam & Adam & Adam \\
Weight decay & 1e-4 & 1e-4 & 1e-4 & 1e-4 & 1e-4 \\
Initial learning rate & 8e-2 & 8e-2 & 8e-2 & 8e-2 & 8e-2 \\
Minimum learning rate & 2e-4 & 2e-4 & 2e-4 & 2e-4 & 2e-4 \\
Epochs & 1000 & 1000 & 150 & 150 & 1000 \\
\bottomrule
\end{tabular}}
\end{table*}

For all experiments, we adopt a residual Forward-Forward (FF) architecture~\cite{sun2025deeperforward} consisting of four residual blocks.
Table~\ref{tab:arch_hparams} summarizes the full architectural configurations and training hyperparameters.
CIFAR-10 and CIFAR-100 use channel widths of $[100, 200, 400, 800]$ across the four residual blocks; MNIST and F-MNIST use lighter widths of $[40, 80, 160, 320]$; and Tiny-ImageNet employs a larger configuration of $[200, 400, 800, 1600]$. The channel width of the very first layer of the network is set equal to the width of the first residual block. Note that the channel dimension must be greater than or equal to the number of classes. Therefore, for Tiny-ImageNet, which contains 200 classes, the model begins with 200 channels. 
For contrastive learning, we employ a projection head implemented by a single linear layer that maps features into a 128-dimensional embedding space.

All models are trained using the Adam optimizer with a weight decay of $1\times10^{-4}$. 
For datasets with larger label spaces, such as CIFAR-100 and Tiny-ImageNet, we use a batch size of 512, while a batch size of 128 is used for CIFAR-10, MNIST, and F-MNIST. 
Models are trained for 1000 epochs on CIFAR-10/100 and Tiny-ImageNet, and for 150 epochs on MNIST and F-MNIST due to their simpler visual complexity. 
We employ cosine annealing from an initial learning rate of $8\times 10^{-2}$ down to $2\times 10^{-4}$.

To stabilize the early stages of contrastive learning and progressively tighten semantic alignment, we schedule the contrastive temperature~$\tau$ throughout training. 
Specifically, $\tau$ is linearly warmed up from $0.8$ to $0.2$ over the first 100 epochs, followed by cosine decay to $0.08$ for the remainder of training.
This schedule mitigates unstable gradients at initialization and encourages increasingly fine-grained feature separation as training progresses.

For natural-image datasets (CIFAR-10, CIFAR-100, and Tiny-ImageNet), we apply strong augmentations to increase intra-class variability.
The augmentation pipeline consists of \texttt{RandomResizedCrop} with a scale range of $(0.6, 1.0)$, \texttt{RandomHorizontalFlip} with probability~0.5, \texttt{ColorJitter} (brightness/contrast/saturation = $0.2$ and hue = $0.1$) with probability~0.8, and \texttt{RandomGrayscale} with probability~0.2. For MNIST and F-MNIST, we adopt lightweight geometric augmentations to avoid distorting digit or clothing structures. 
The augmentation pipeline includes \texttt{RandomRotation} within $\pm 10^\circ$ and \texttt{RandomAffine} with up to 10\% translation in both spatial directions and a scaling factor in the range $[0.9, 1.1]$.

All HCL-FF experiments follow a two-stage training procedure. In the first stage, we pretrain a model without hierarchical learning. The resulting pretrained model is then used to construct the data-driven hierarchy. In the second stage, we train the full HCL-FF model from scratch with both hierarchical and contrastive learning.

For comparison, the BP-trained ResNet-20 baseline is reproduced using the exact configuration described in the original ResNet paper~\cite{he2016deep}. 
We also train a matching-parameter variant by increasing all channel widths of the original configuration by a factor of ten to closely match the parameter count of our HCL-FF models.

\section{Residual Shortcuts}
\label{supp::residual}

\begin{table}[t]
\centering
\small
\setlength{\tabcolsep}{4pt}
\caption{Effect of residual shortcuts on accuracy.}
\resizebox{0.8\columnwidth}{!}{
\begin{tabular}{lcc}
\toprule
\textbf{Dataset} & \textbf{w/o residual} & \textbf{w/ residual} \\
\midrule
CIFAR-10 Accuracy & 90.85 & 91.83 \\
CIFAR-100 Accuracy & 65.43 & 70.76 \\
\bottomrule
\end{tabular}
}
\label{tab:residual}
\end{table}

Residual structures are commonly used in backpropagation-based (BP) networks to ease optimization by providing gradient shortcut pathways~\cite{he2016deep}.
In the FF setting, where gradients are not propagated across layers, residual shortcuts instead serve to fuse information across depths.
Following DeeperForward~\cite{sun2025deeperforward}, we adopt parameter-free shortcuts that adapt spatial and channel dimensions without introducing any learnable projection layers.
For completeness, we describe the residual mechanism in detail in this section.

\paragraph{Spatial alignment.}
Let $\mathbf{z}^{(\ell)}_{r} \in \mathbb{R}^{C \times H_r \times W_r}$ denote the shortcut feature for layer $\ell$, and let the goodness-decoupled feature of the current layer be $\mathbf{z}^{(\ell)} \in \mathbb{R}^{C \times H \times W}$.
When transitioning between residual blocks, the spatial resolution is downsampled by a factor of two, causing $(H_r, W_r)\neq(H, W)$.
We therefore apply a $2\times 2$ average-pooling operation with stride~2 when a resolution mismatch occurs:
\begin{equation}
\mathbf{\tilde{z}}^{(\ell)}_{r} =
\begin{cases}
\text{AvgPool}_{2\times2}\!\left(\mathbf{z}^{(\ell)}_r\right), & (H_r, W_r) \neq (H, W), \\[2pt]
\mathbf{z}^{(\ell)}_r, & (H_r, W_r) = (H, W).
\end{cases}
\label{eq:supp_residual_pool}
\end{equation}

\paragraph{Channel alignment across residual blocks.}
At the transition between residual blocks, the main branch doubles its channel width.
Let $\mathbf{z}^{(\ell)} \in \mathbb{R}^{C \times H \times W}$ be the goodness-decoupled feature at the end of one residual block, and let the shortcut feature be $\mathbf{\tilde{z}}^{(\ell)}_{r} \in \mathbb{R}^{C \times H \times W}$.
To construct the input to the next residual block, which expects $2C$ channels, the two tensors are concatenated along the channel dimension:
\begin{equation}
\mathbf{z}^{(\ell)}_{\text{merge}}
= 
\text{Concat}_{\text{channel}}\!\left(\mathbf{z}^{(\ell)},\ \mathbf{\tilde{z}}^{(\ell)}_{r}\right)
\in \mathbb{R}^{(2C) \times H \times W}.
\label{eq:supp_residual_concat}
\end{equation}
This yields a channel-aligned shortcut without any learned projection.

\paragraph{Residual fusion within a residual block.}
Within the same residual block, the main branch and shortcut share the same spatial and channel dimensions.
Thus, we perform simple additive fusion:
\begin{equation}
\mathbf{z}^{(\ell)}_{\text{merge}} = \mathbf{z}^{(\ell)} + \mathbf{\tilde{z}}^{(\ell)}_{r}
\in \mathbb{R}^{C\times H\times W}.
\label{eq:supp_residual_add}
\end{equation}

\paragraph{Discussion.}
The shortcut path participates in neither gradient flow nor weight updates, preserving the locality of FF training.
However, these lightweight residuals substantially improve representational quality by propagating useful features across depth.
As shown in Table~\ref{tab:residual}, removing residual shortcuts causes significant accuracy degradation on CIFAR-10 and CIFAR-100, highlighting their importance for stable and effective layer-wise FF training.

\section{Signal Integrating and Pruning Module}
\label{supp::sip}

In this section, we elaborate on the Signal Integrating and Pruning (SIP) module introduced in DeeperForward~\cite{sun2025deeperforward}. 
Unlike BP-based models, which rely solely on the final layer’s logits, FF networks produce a goodness score at every layer.
Each layer-wise goodness vector $\mathbf{g}^{(\ell)}$ already encodes partial class evidence at depth~$\ell$, providing an opportunity to aggregate predictions across the entire network hierarchy.

\paragraph{Formulation.}
Given a model with $L$ layers, each layer produces a class-wise goodness vector $\mathbf{g}^{(\ell)} \in \mathbb{R}^{K}$, where $K$ is the number of classes. 
Instead of relying exclusively on the deepest layer, SIP forms an aggregated prediction by averaging goodness scores over a selected contiguous interval of layers:
\begin{equation}
\tilde{\mathbf{g}}
=
\frac{1}{e-s+1}
\sum_{\ell = s}^{e}
{\mathbf{g}}^{(\ell)},
\qquad
0 \leq s \leq e \leq L-1.
\label{eq:supp_sip_sum}
\end{equation}
The final class label is obtained by taking the $\arg\max$ over $\tilde{\mathbf{g}}$. The interval $[s, e]$ is chosen on a validation set and kept fixed during inference.

\begin{table}[t]
\centering
\small
\setlength{\tabcolsep}{4pt}
\caption{Accuracy under different prediction strategies. 
"All layers" averages goodness scores from all layers, 
"Last layer" uses only the deepest layer, 
and "SIP" reports the accuracy and the selected interval $[s, e]$ found on the validation set.}
\resizebox{\columnwidth}{!}{
\begin{tabular}{lccc}
\toprule
\textbf{Method} & \textbf{All layers} & \textbf{Last layer} & \textbf{SIP [s, e]} \\
\midrule
\multicolumn{4}{l}{\textbf{MNIST}} \\
\quad DeeperForward   & 99.61 & 99.50 & 99.67 [4, 8] \\
\quad Ours            & 99.51 & 99.62 & 99.70 [13, 16] \\
\midrule
\multicolumn{4}{l}{\textbf{F-MNIST}} \\
\quad DeeperForward   & 93.01 & 91.73 & 92.97 [1, 13] \\
\quad Ours            & 89.15 & 93.75 & 93.97 [13, 16] \\
\midrule
\multicolumn{4}{l}{\textbf{CIFAR-10}} \\
\quad DeeperForward   & 86.16 & 79.77 & 86.24 [1, 16] \\
\quad Ours            & 80.39 & 92.00 & 91.83 [10, 16] \\
\midrule
\multicolumn{4}{l}{\textbf{CIFAR-100}} \\
\quad DeeperForward   & 52.63 & 26.56 & 52.43 [0, 12] \\
\quad Ours            & 49.23 & 67.42 & 70.76 [7, 16] \\
\midrule
\multicolumn{4}{l}{\textbf{Tiny-ImageNet}} \\
\quad DeeperForward   & 33.96 & 3.05 & 35.95 [1, 12] \\
\quad Ours            & 33.39 & 48.46 & 48.46 [16, 16] \\
\bottomrule
\end{tabular}
}
\label{tab:sip}
\end{table}

\paragraph{Analysis.}
Table~\ref{tab:sip} reveals distinct SIP behaviors for DeeperForward and HCL-FF. 
For HCL-FF, SIP consistently selects deeper intervals, e.g., $[13,16]$ on MNIST and F-MNIST, $[10,16]$ on CIFAR-10, $[7,16]$ on CIFAR-100, and $[16,16]$ on Tiny-ImageNet.
This pattern aligns with our hierarchical design: early layers focus on coarse distinctions, while deeper layers progressively refine semantics, making them more informative for final prediction.
In contrast, DeeperForward frequently selects shallow or mid-level layers, indicating that deeper layers contribute weaker or noisier predictions. This supports our analysis that goodness decoupling, without hierarchical or contrastive grounding, leads to semantic drift in deeper representations, preventing meaningful development across depth.

We additionally consider two baselines: averaging goodness across all layers and using only the final layer.
For HCL-FF, all-layer averaging reduces performance because early layers produce coarse super-class signals that hinder fine-grained classification.
However, our last-layer accuracy substantially exceeds that of DeeperForward, demonstrating that HCL-FF maintains discriminative structure even at the deepest layers and benefits from a meaningful depth-wise learning trajectory.

\section{Building Hierarchies}
\label{supp::building}

In this section, we describe how we construct the class hierarchy used for hierarchical supervision in detail and clarify the rationale behind this design. 
Following the Neural-Backed Decision Tree (NBDT) framework~\cite{wan2020nbdt}, we employ a data-driven procedure that derives a semantic hierarchy directly from a trained classifier. 
This approach eliminates the need for manual specification of coarse and fine semantic groups, offering a systematic method to construct a hierarchy for any dataset.

\paragraph{Classifier weights as class prototypes.}
As outlined in Sec.~3.2 of the main paper, we begin by training a linear classifier on top of the frozen backbone.
Let $W \in \mathbb{R}^{K \times D}$ denote the weight matrix of a trained linear classifier over $K$ classes, where each row $W_{k} \in \mathbb{R}^{D}$ corresponds to class $k$.  
For an input feature $\mathbf{z}$, the classifier logit for class~$k$ is
\begin{equation}
\text{logit}_{k}(\mathbf{z}) 
= W_{k}^{\top}\mathbf{z} + b_{k}.
\end{equation}  
Because the dot product $W_{k}^{\top}\mathbf{z}$ is maximized when $\mathbf{z}$ aligns with $W_{k}$, each row vector $W_{k}$ implicitly acts as the prototype or anchor direction defining the region of feature space associated with class $k$.  
This interpretation is well supported by prior work~\cite{wang2017normface, liu2017sphereface, wang2018cosface, wan2020nbdt}. 

\paragraph{From prototypes to semantic hierarchy.}
Before computing inter-class similarity, each prototype is L2-normalized,
$\hat{W}_{k} = \frac{W_{k}}{\|W_{k}\|_{2}}$.
We then compute pairwise cosine distances,
\begin{equation}
d(k, k') = 1 - \hat{W}_{k}^{\top} \hat{W}_{k'},
\end{equation}
and apply hierarchical agglomerative clustering with Ward linkage to recursively merge the most similar classes.  
This produces a full binary hierarchy that reflects the semantic relationships encoded by the classifier.  
Intuitively, classes with similar prototype directions are more confusable and therefore merge earlier in the hierarchy, yielding a semantically meaningful tree structure.

\begin{figure}[t]
    \centering
    \includegraphics[width=\columnwidth]{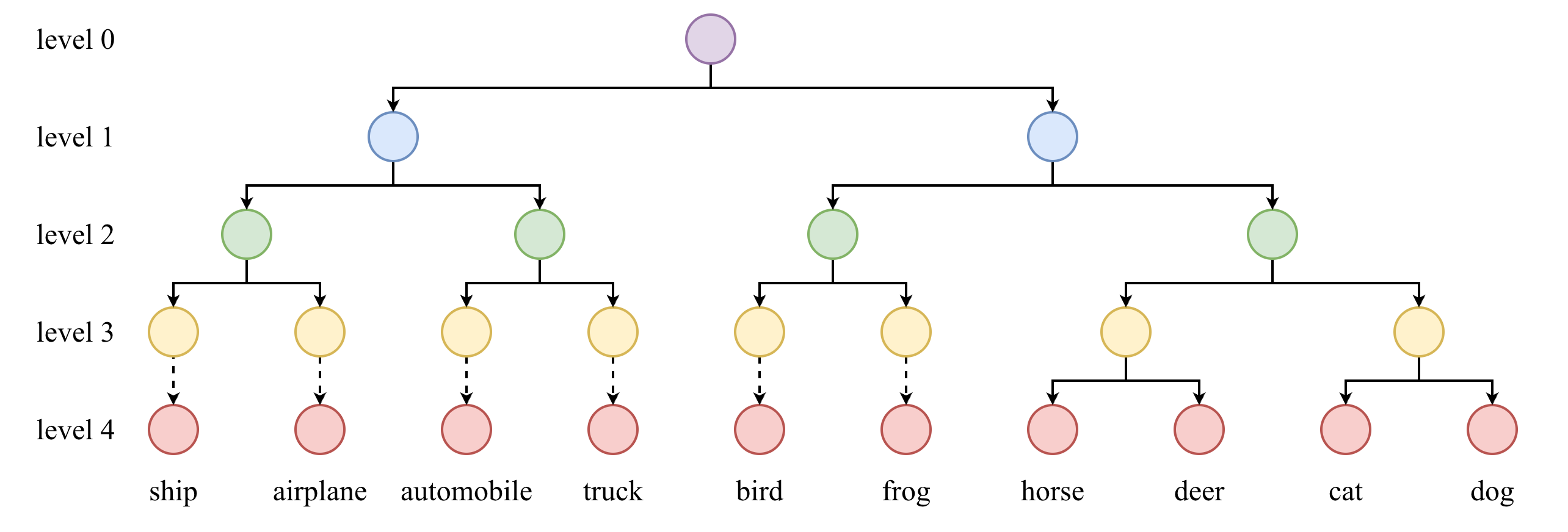}
    \caption{
    Hierarchy of CIFAR-10 classes constructed via clustering on class prototypes derived from a pre-trained classifier.
    Each level represents a different granularity of super-class grouping.
    }
    \label{fig:hierarchy_supp}
\end{figure}

\section{Super-class Accuracy}
\label{supp::super}
\begin{figure}[t]
    \centering
    \includegraphics[width=\columnwidth]{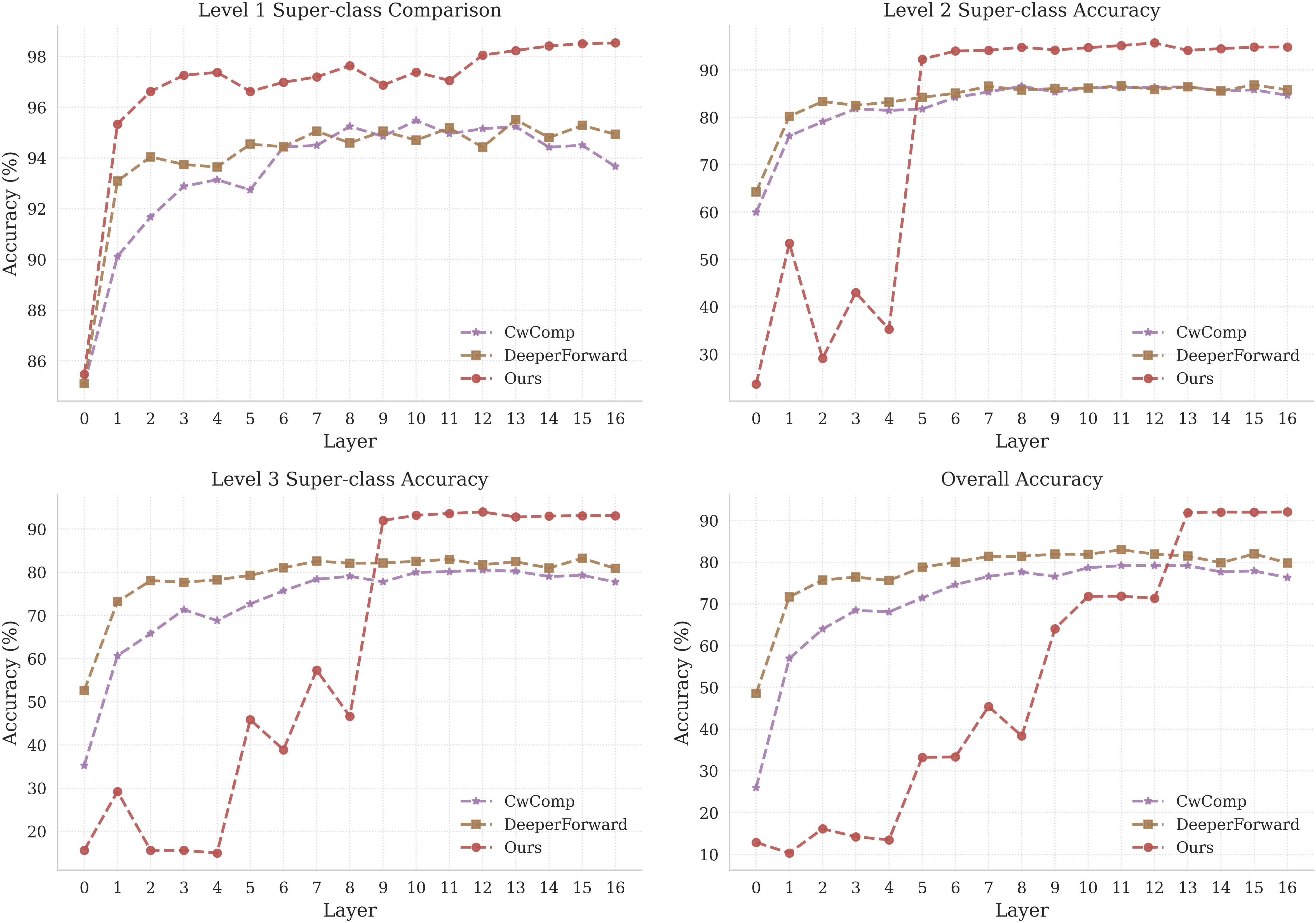}
    \caption{
    Layer-wise super-class accuracy at different hierarchy levels and overall fine-grained accuracy.
    }
    \label{fig:layer_acc_2x2}
\end{figure}

To complement the per-layer analysis presented in Fig.~3 of the main paper, we further examine layer-wise accuracy at each hierarchy level, comparing against CwComp~\cite{papachristodoulou2024convolutional} and DeeperForward~\cite{sun2025deeperforward}. 
Figure~\ref{fig:layer_acc_2x2} reports Level~1, Level~2, and Level~3 super-class accuracy, as well as overall fine-grained accuracy. 
The hierarchy levels correspond to those defined in Fig.~\ref{fig:hierarchy_supp}.

Our model exhibits a clear and structured progression across depth. 
In the shallow network from layer 0 to layer 4, HCL-FF achieves strong performance on Level~1 super-classes while showing limited ability to discriminate fine-grained categories, which is precisely the behavior encouraged by hierarchical supervision. 
As depth increases, the model receives finer-grained targets and correspondingly improves its performance on deeper hierarchy levels and on fine-grained classification. This demonstrates that hierarchical supervision successfully guides the network to develop increasingly specialized semantic representations.

In contrast, both CwComp and DeeperForward require the shallow layers to immediately solve the full fine-grained classification problem. 
Although this can yield higher fine-grained accuracy in the early layers, it comes at the expense of poorer performance on coarse super-class distinctions.
This pattern suggests that these methods prematurely force early layers toward high-level abstraction, reducing the stability and generality of their representations.

\section{Model Efficiency}
\label{supp::efficiency}

\begin{figure}[t]
    \centering
    \includegraphics[width=\columnwidth]{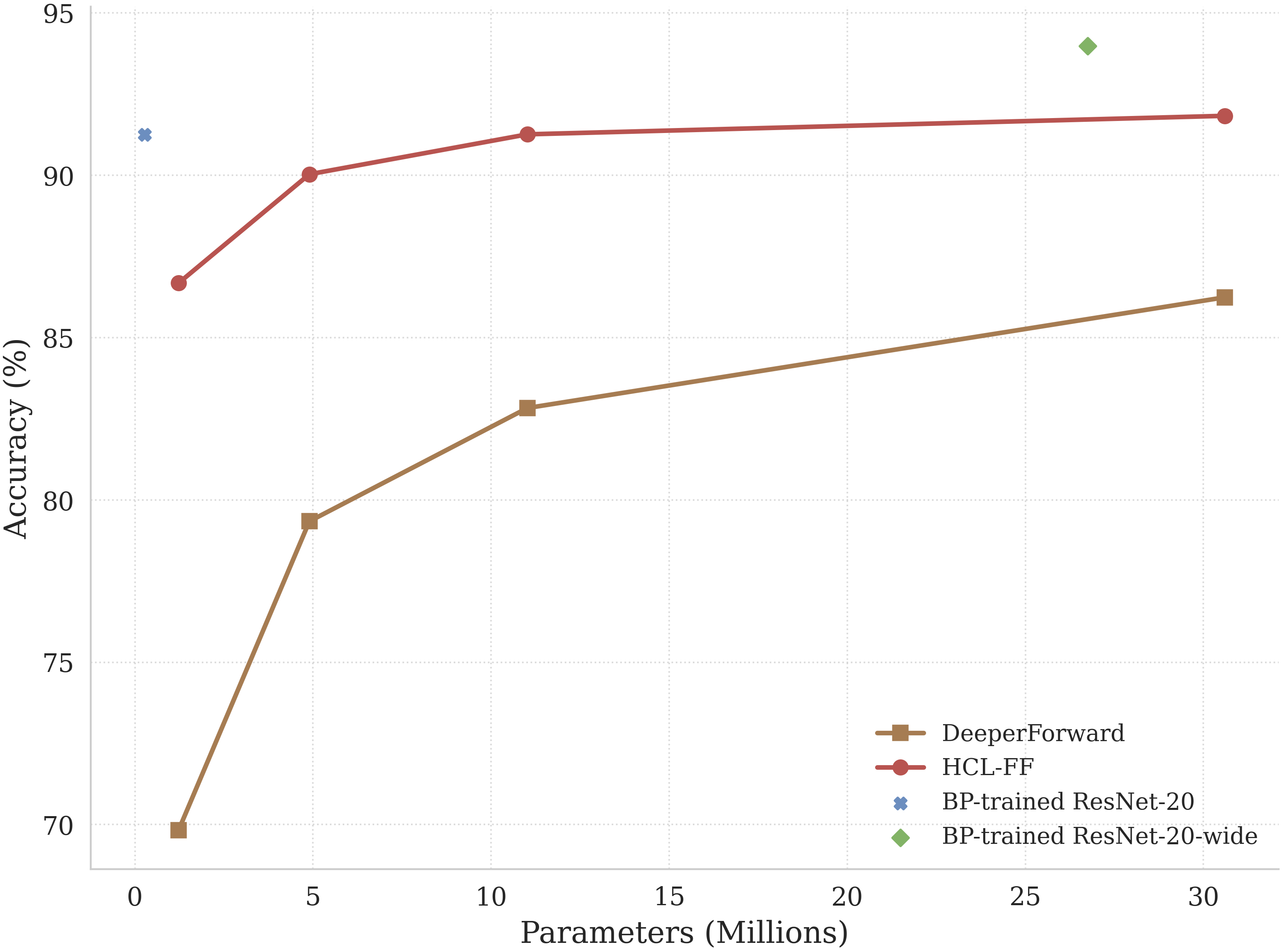}
    \caption{
    CIFAR-10 accuracy versus parameter count for DeeperForward, HCL-FF, and BP-trained ResNet baselines.
    }
    \label{fig:param_efficiency}
\end{figure}

To evaluate parameter efficiency, we compare HCL-FF with DeeperForward, along with BP-trained ResNet-20 and the parameter-matched ResNet-20-wide baseline.
Figure~\ref{fig:param_efficiency} reports CIFAR-10 accuracy as a function of total parameter count. 
We evaluate four model capacities for both HCL-FF and DeeperForward, where the channel widths of the four residual blocks are set as follows:
\begin{itemize}
\item \textbf{Tiny}: $[20, 40, 80, 160]$ channels,
\item \textbf{Small}: $[40, 80, 160, 320]$ channels,
\item \textbf{Medium}: $[60, 120, 240, 480]$ channels,
\item \textbf{Large}: $[100, 200, 400, 800]$ channels (as in Sec.~\ref{supp::arch}).
\end{itemize}

Across all model sizes, HCL-FF achieves substantially higher accuracy than DeeperForward.
The HCL-FF curve consistently lies above that of DeeperForward, demonstrating a markedly more favorable accuracy-capacity trade-off.
Moreover, the tiny configuration (1.225M parameters) of our HCL-FF model attains 86.68\% accuracy, exceeding DeeperForward’s large model (30.603M parameters, 86.24\%) while using less than 5\% of its parameters.

Nevertheless, we acknowledge that a performance gap remains between layer-wise training and full end-to-end BP optimization. 
Our medium configuration (11.018M parameters) achieves a comparable accuracy of 91.26\% to BP-trained ResNet-20 (91.25\%). However, it requires more parameters to achieve this level of accuracy.
Moreover, our large HCL-FF model still falls short of the parameter-matched ResNet-20-wide by a few percentage points.
This suggests that while hierarchical and contrastive guidance greatly enhance FF-based training, further advances in local learning objectives are needed to close the remaining gap to BP.

Overall, the results in Figure~\ref{fig:param_efficiency} show that HCL-FF delivers significantly improved parameter efficiency over prior FF-based methods, bringing FF training meaningfully closer to BP-trained networks while maintaining its fully local training paradigm.

\section{Learning Dynamics}
\label{supp::learning}
\begin{figure}[t]
\centering
\includegraphics[width=\columnwidth]{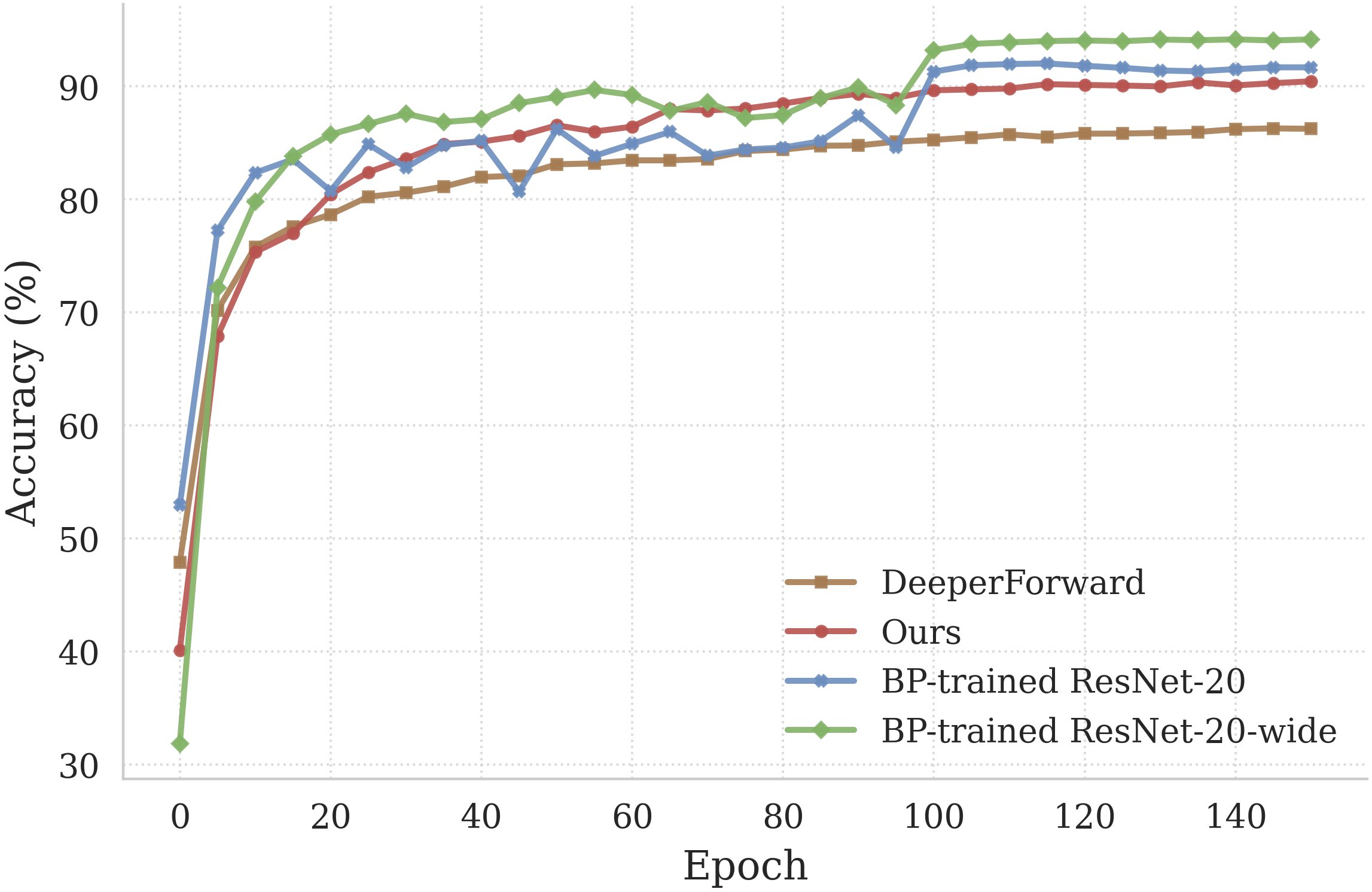}
\caption{
CIFAR-10 training curves under a matched 150-epoch schedule.
HCL-FF converges faster and reaches higher accuracy than DeeperForward.}
\label{fig:learning_curve}
\end{figure}

To further compare the convergence behavior of HCL-FF, DeeperForward, and standard backpropagation, we examine their CIFAR-10 learning curves under a matched training schedule.
All models are trained for 150 epochs for a fair comparison.
Figure~\ref{fig:learning_curve} reports test accuracy across training epochs.

HCL-FF converges noticeably faster and to a higher accuracy than DeeperForward.
By epoch~20, HCL-FF already surpasses DeeperForward and maintains this advantage throughout training, indicating that hierarchical supervision and contrastive grounding provide stronger and more stable learning signals.

BP-trained ResNet-20 exhibits larger fluctuations during the early and mid stages of training, reflecting the dynamics of joint gradient updates across all layers.
Despite this instability, it eventually achieves higher final accuracy, highlighting the remaining challenges of purely layer-wise optimization compared to end-to-end backpropagation.

Overall, the learning curves confirm that HCL-FF delivers both faster and smoother convergence than prior FF-based approaches, while narrowing the gap to BP-trained networks under identical training conditions.

\section{Hierarchy Mapping}
\label{supp::mapping}

\begin{table}[t]
\centering
\small
\setlength{\tabcolsep}{4pt}
\caption{Effect of alternative hierarchy mapping strategies.}
\resizebox{\columnwidth}{!}{
\begin{tabular}{lccc}
\toprule
Dataset & Incremental & Decremental & Balanced \\
\midrule
CIFAR-100 Accuracy & 70.30 & 68.40 & 70.76 \\
\bottomrule
\end{tabular}
}
\label{tab:mapping}
\end{table}

In addition to the hierarchy mapping method described in Eq.~10 of the main paper, we evaluate two alternative strategies for assigning hierarchy levels across layers.

\paragraph{(1) Incremental mapping.}
This strategy assigns coarse supervision to the earliest layers and gradually increases granularity with depth.
Specifically, for a network with $L$ layers and a hierarchy of depth $D$, we define:
\begin{align}
\text{level}(i) = \min(1+i, D), \quad i = 0, \dots, L-1,
\end{align}
Thus, layer~0 receives Level~1 supervision, layer~1 receives Level~2, and so forth, until the maximum depth $D$ is reached, after which all deeper layers are supervised at Level~$D$.
This mapping encourages early layers to transition quickly toward finer semantic resolutions.

\paragraph{(2) Decremental mapping.}
Conversely, this strategy assigns fine-grained supervision to the deepest layers and progressively coarser supervision toward the input:
\begin{align}
\text{level}(i) = \max(D-(L-1-i), 1), \quad i = 0, \dots, L-1,
\end{align}
Here, shallow and mid-level layers are trained on very coarse super-classes, and only the final few layers receive fine-grained targets.

\paragraph{Results.}
Table~\ref{tab:mapping} compares the performance of these mapping strategies.
For clarity, we refer to the hierarchy assignment defined in Eq.~10 of the main paper as the \emph{balanced} mapping, since it distributes hierarchy levels more evenly across depth.
The incremental mapping performs similarly to the balanced strategy, while the decremental mapping results in a clear performance drop.
This outcome reflects the importance of assigning sufficiently fine supervision to the middle layers. Although these layers have substantial representational capacity, the decremental mapping constrains them to overly coarse tasks, preventing the model from developing nuanced intermediate features.
As a result, fine-grained supervision is concentrated only at the very end of the network, and the deeper hierarchy levels are not effectively leveraged across depth.

% \section{Hierarchy Robustness}
% \label{supp::hier_robust}

\section{Effect of Batch Size}
\label{supp::bs}

\begin{table}[t]
\centering
\small
\setlength{\tabcolsep}{4pt}
\caption{Effect of batch size.}
\resizebox{\columnwidth}{!}{
\begin{tabular}{lccccc}
\toprule
Batch size & 32 & 64 & 128 & 256 & 512 \\
\midrule
CIFAR-10 Accuracy & 92.10 & 92.22 & 91.83 & 91.73 & 91.96 \\
\bottomrule
\end{tabular}
}
\label{tab:bs}
\end{table}

Table~\ref{tab:bs} shows the effect of batch size on CIFAR-10 performance.
HCL-FF demonstrates stable performance across a broad range, with accuracy varying by less than 0.5\% between batch sizes of 32 and 512.
This indicates that the contrastive component of HCL-FF does not rely on extremely large batches, in contrast to many contrastive learning frameworks that require thousands of samples per batch for stable gradients.
Two factors may contribute to this robustness.
First, layer-wise training restricts each update to a smaller subset of parameters, which stabilizes optimization and reduces the need for large batch sizes.
Second, hierarchical supervision provides a strong coarse-to-fine semantic structure that helps maintain stable training dynamics even at smaller batch scales.

Finally, we observe slightly higher accuracy at smaller batch sizes.
Because the number of epochs is fixed, smaller batches yield more update steps.
Since layer-wise learning is less sample-efficient than end-to-end backpropagation, these additional updates give deeper layers more opportunities to refine their representations using information propagated from earlier layers.

\begin{table*}[t]
\centering
\small
\resizebox{\textwidth}{!}{
\begin{tabular}{lcccc|c}
\toprule
\textbf{Method} & \textbf{ResNet20-Wide} & \textbf{DeeperForward} & \textbf{Ours-pretraining} & \textbf{Ours} & \textbf{DeeperForward-extend} \\
\midrule
CIFAR-100 Acc. & 74.91 & 53.57 & 63.63 & 69.03 & 55.01 \\
Training GPU Hours (NVIDIA RTX A6000) & 2.21 & 0.79 & 1.32 & 1.36 & 3.02 \\
Training GPU Memory Usage (MiB) & 9916 & 3552 & 3848 & 3848 & 3552 \\
\bottomrule
\end{tabular}
}
\caption{Comparison of computational cost and performance on CIFAR-100.}
\label{tab:computation}
\end{table*}

\section{Computational Cost}
\label{supp::cost}

To demonstrate the computational aspect of HCL-FF, we compare HCL-FF against a BP-trained ResNet variant and DeeperForward under matched training conditions.
All models have comparable numbers of parameters and are trained for 200 epochs with a batch size of 512, ensuring fair model size and equal numbers of weight updates.

For HCL-FF, we report the computational cost of its two training stages separately.
The first stage (\emph{Ours-pretraining}) trains the model without hierarchical learning and is used to construct the data-driven hierarchy.
The second stage (\emph{Ours}) trains the full model from scratch with both hierarchical and contrastive objectives.
To further contextualize the computational overhead, we additionally include an extended DeeperForward baseline trained for 750 epochs (\emph{DeeperForward-extend}), which approximately matches the total GPU hours consumed by our two-stage pipeline.

Table~\ref{tab:computation} summarizes classification accuracy, total GPU hours, and GPU memory usage.
Several observations can be made. First, FF-based methods are substantially more memory-efficient than BP-based training. ResNet20-Wide requires 9916\,MiB of GPU memory, whereas all FF-based variants require between 3552 and 3848\,MiB. This reduction stems from the layer-wise local learning paradigm, which avoids storing full-network activations for backpropagation.

Second, incorporating contrastive learning introduces only a modest increase in memory usage compared to DeeperForward (3848\,MiB vs.\ 3552\,MiB), while providing a significant improvement in accuracy (63.63\% vs.\ 53.57\%).
Although the contrastive objective increases total GPU hours, the computational overhead remains moderate relative to the accuracy gains.

Third, the additional cost of hierarchical learning is minimal.
The final HCL-FF model (\emph{Ours}) exhibits nearly identical GPU memory usage and training time to the pretraining stage (\emph{Ours-pretraining}), while achieving a further improvement in accuracy (69.03\% vs.\ 63.63\%).
This indicates that hierarchical learning introduces negligible computational overhead while achieving substantial accuracy gains.

Finally, increasing computation alone does not bridge the performance gap.
The extended DeeperForward baseline consumes more GPU hours (3.02 hours) than our complete two-stage pipeline (1.32 + 1.36 hours), yet achieves substantially lower accuracy (55.01\% vs.\ 69.03\%).
This demonstrates that the performance gains of HCL-FF arise from the proposed hierarchical and contrastive learning mechanisms rather than from increased computational budget.

Overall, HCL-FF achieves significant accuracy improvements over prior FF-based methods with only modest additional training time and minimal memory overhead, while retaining the memory efficiency advantages of fully local learning.

%% file: main.bib
@String(AAAI = {AAAI})

@article{hinton2022forward,
  title={The forward-forward algorithm: Some preliminary investigations},
  author={Hinton, Geoffrey},
  journal={arXiv preprint arXiv:2212.13345},
  volume={2},
  number={3},
  pages={5},
  year={2022}
}

@article{dooms2023trifecta,
  title={The trifecta: Three simple techniques for training deeper forward-forward networks},
  author={Dooms, Thomas and Tsang, Ing Jyh and Oramas, Jose},
  journal={arXiv preprint arXiv:2311.18130},
  year={2023}
}

@article{lee2023symba,
  title={Symba: Symmetric backpropagation-free contrastive learning with forward-forward algorithm for optimizing convergence},
  author={Lee, Heung-Chang and Song, Jeonggeun},
  journal={arXiv preprint arXiv:2303.08418},
  year={2023}
}

@inproceedings{papachristodoulou2024convolutional,
  title={Convolutional channel-wise competitive learning for the forward-forward algorithm},
  author={Papachristodoulou, Andreas and Kyrkou, Christos and Timotheou, Stelios and Theocharides, Theocharis},
  booktitle={Proceedings of the AAAI Conference on Artificial Intelligence},
  volume={38},
  number={13},
  pages={14536--14544},
  year={2024}
}

@article{zhao2025cascaded,
  title={The cascaded forward algorithm for neural network training},
  author={Zhao, Gongpei and Wang, Tao and Jin, Yi and Lang, Congyan and Li, Yidong and Ling, Haibin},
  journal={Pattern Recognition},
  volume={161},
  pages={111292},
  year={2025},
  publisher={Elsevier}
}

@inproceedings{sun2025deeperforward,
  title={DeeperForward: Enhanced Forward-Forward Training for Deeper and Better Performance},
  author={Sun, Liang and Zhang, Yang and Wen, Jiajun and Shen, Linlin and Xie, Weicheng and He, Weizhao},
  booktitle={International Conference on Learning Representations},
  year={2025}
}

@inproceedings{he2016deep,
  title={Deep residual learning for image recognition},
  author={He, Kaiming and Zhang, Xiangyu and Ren, Shaoqing and Sun, Jian},
  booktitle={Proceedings of the IEEE conference on computer vision and pattern recognition},
  pages={770--778},
  year={2016}
}

@article{wan2020nbdt,
  title={NBDT: Neural-backed decision trees},
  author={Wan, Alvin and Dunlap, Lisa and Ho, Daniel and Yin, Jihan and Lee, Scott and Jin, Henry and Petryk, Suzanne and Bargal, Sarah Adel and Gonzalez, Joseph E},
  journal={arXiv preprint arXiv:2004.00221},
  year={2020}
}

@inproceedings{Le2015TinyIV,
  title={Tiny ImageNet Visual Recognition Challenge},
  author={Ya Le and Xuan S. Yang},
  year={2015},
  url={https://api.semanticscholar.org/CorpusID:16664790}
}

@article{krizhevsky2009learning,
  title={Learning multiple layers of features from tiny images},
  author={Krizhevsky, Alex and Hinton, Geoffrey and others},
  year={2009},
  publisher={Toronto, ON, Canada}
}

@article{chen2025self,
  title={Self-Contrastive Forward-Forward Algorithm},
  author={Chen, Xing and Liu, Dongshu and Laydevant, J{\'e}r{\'e}mie and Grollier, Julie},
  journal={Nature Communications},
  volume={16},
  number={1},
  pages={5978},
  year={2025},
  publisher={Nature Publishing Group UK London}
}

@article{srinivasan2023forward,
  title={Forward learning with top-down feedback: Empirical and analytical characterization},
  author={Srinivasan, Ravi and Mignacco, Francesca and Sorbaro, Martino and Refinetti, Maria and Cooper, Avi and Kreiman, Gabriel and Dellaferrera, Giorgia},
  journal={arXiv preprint arXiv:2302.05440},
  year={2023}
}

@inproceedings{ororbia2023backpropagation,
  title={Backpropagation-free deep learning with recursive local representation alignment},
  author={Ororbia, Alexander G and Mali, Ankur and Kifer, Daniel and Giles, C Lee},
  booktitle={Proceedings of the AAAI conference on artificial intelligence},
  volume={37},
  number={8},
  pages={9327--9335},
  year={2023}
}

@article{kohan2023signal,
  title={Signal propagation: The framework for learning and inference in a forward pass},
  author={Kohan, Adam and Rietman, Edward A and Siegelmann, Hava T},
  journal={IEEE Transactions on Neural Networks and Learning Systems},
  volume={35},
  number={6},
  pages={8585--8596},
  year={2023},
  publisher={IEEE}
}

@article{journe2022hebbian,
  title={Hebbian deep learning without feedback},
  author={Journ{\'e}, Adrien and Rodriguez, Hector Garcia and Guo, Qinghai and Moraitis, Timoleon},
  journal={arXiv preprint arXiv:2209.11883},
  year={2022}
}

@inproceedings{ernoult2022towards,
  title={Towards scaling difference target propagation by learning backprop targets},
  author={Ernoult, Maxence M and Normandin, Fabrice and Moudgil, Abhinav and Spinney, Sean and Belilovsky, Eugene and Rish, Irina and Richards, Blake and Bengio, Yoshua},
  booktitle={International Conference on Machine Learning},
  pages={5968--5987},
  year={2022},
  organization={PMLR}
}

@article{rumelhart1986learning,
  title={Learning representations by back-propagating errors},
  author={Rumelhart, David E and Hinton, Geoffrey E and Williams, Ronald J},
  journal={nature},
  volume={323},
  number={6088},
  pages={533--536},
  year={1986},
  publisher={Nature Publishing Group UK London}
}

@article{crick1989recent,
  title={The recent excitement about neural networks},
  author={Crick, Francis},
  journal={Nature},
  volume={337},
  number={6203},
  pages={129--132},
  year={1989},
  publisher={Nature Publishing Group UK London}
}

@article{grossberg1987competitive,
  title={Competitive learning: From interactive activation to adaptive resonance},
  author={Grossberg, Stephen},
  journal={Cognitive science},
  volume={11},
  number={1},
  pages={23--63},
  year={1987},
  publisher={Elsevier}
}

@article{khosla2020supervised,
  title={Supervised contrastive learning},
  author={Khosla, Prannay and Teterwak, Piotr and Wang, Chen and Sarna, Aaron and Tian, Yonglong and Isola, Phillip and Maschinot, Aaron and Liu, Ce and Krishnan, Dilip},
  journal={Advances in neural information processing systems},
  volume={33},
  pages={18661--18673},
  year={2020}
}

@inproceedings{lee2015difference,
  title={Difference target propagation},
  author={Lee, Dong-Hyun and Zhang, Saizheng and Fischer, Asja and Bengio, Yoshua},
  booktitle={Joint european conference on machine learning and knowledge discovery in databases},
  pages={498--515},
  year={2015},
  organization={Springer}
}

@article{lillicrap2016random,
  title={Random synaptic feedback weights support error backpropagation for deep learning},
  author={Lillicrap, Timothy P and Cownden, Daniel and Tweed, Douglas B and Akerman, Colin J},
  journal={Nature communications},
  volume={7},
  number={1},
  pages={13276},
  year={2016},
  publisher={Nature Publishing Group UK London}
}

@article{nokland2016direct,
  title={Direct feedback alignment provides learning in deep neural networks},
  author={N{\o}kland, Arild},
  journal={Advances in neural information processing systems},
  volume={29},
  year={2016}
}

@article{akrout2019deep,
  title={Deep learning without weight transport},
  author={Akrout, Mohamed and Wilson, Collin and Humphreys, Peter and Lillicrap, Timothy and Tweed, Douglas B},
  journal={Advances in neural information processing systems},
  volume={32},
  year={2019}
}

@article{bartunov2018assessing,
  title={Assessing the scalability of biologically-motivated deep learning algorithms and architectures},
  author={Bartunov, Sergey and Santoro, Adam and Richards, Blake and Marris, Luke and Hinton, Geoffrey E and Lillicrap, Timothy},
  journal={Advances in neural information processing systems},
  volume={31},
  year={2018}
}

@article{bengio2014auto,
  title={How auto-encoders could provide credit assignment in deep networks via target propagation},
  author={Bengio, Yoshua},
  journal={arXiv preprint arXiv:1407.7906},
  year={2014}
}

@article{frenkel2021learning,
  title={Learning without feedback: Fixed random learning signals allow for feedforward training of deep neural networks},
  author={Frenkel, Charlotte and Lefebvre, Martin and Bol, David},
  journal={Frontiers in neuroscience},
  volume={15},
  pages={629892},
  year={2021},
  publisher={Frontiers Media SA}
}

@article{ororbia2023brain,
  title={Brain-inspired machine intelligence: A survey of neurobiologically-plausible credit assignment},
  author={Ororbia, Alexander G},
  journal={arXiv preprint arXiv:2312.09257},
  year={2023}
}

@article{launay2019principled,
  title={Principled training of neural networks with direct feedback alignment},
  author={Launay, Julien and Poli, Iacopo and Krzakala, Florent},
  journal={arXiv preprint arXiv:1906.04554},
  year={2019}
}

@article{moraitis2022softhebb,
  title={Softhebb: Bayesian inference in unsupervised hebbian soft winner-take-all networks},
  author={Moraitis, Timoleon and Toichkin, Dmitry and Journ{\'e}, Adrien and Chua, Yansong and Guo, Qinghai},
  journal={Neuromorphic Computing and Engineering},
  volume={2},
  number={4},
  pages={044017},
  year={2022},
  publisher={IOP Publishing}
}

@article{rumelhart1985feature,
  title={Feature discovery by competitive learning},
  author={Rumelhart, David E and Zipser, David},
  journal={Cognitive science},
  volume={9},
  number={1},
  pages={75--112},
  year={1985},
  publisher={Elsevier}
}

@article{song2000competitive,
  title={Competitive Hebbian learning through spike-timing-dependent synaptic plasticity},
  author={Song, Sen and Miller, Kenneth D and Abbott, Larry F},
  journal={Nature neuroscience},
  volume={3},
  number={9},
  pages={919--926},
  year={2000},
  publisher={Nature Publishing Group}
}

@article{miconi2021hebbian,
  title={Hebbian learning with gradients: Hebbian convolutional neural networks with modern deep learning frameworks},
  author={Miconi, Thomas},
  journal={arXiv preprint arXiv:2107.01729},
  year={2021}
}

@article{lagani2021hebbian,
  title={Hebbian semi-supervised learning in a sample efficiency setting},
  author={Lagani, Gabriele and Falchi, Fabrizio and Gennaro, Claudio and Amato, Giuseppe},
  journal={Neural Networks},
  volume={143},
  pages={719--731},
  year={2021},
  publisher={Elsevier}
}

@inproceedings{dellaferrera2022error,
  title={Error-driven input modulation: solving the credit assignment problem without a backward pass},
  author={Dellaferrera, Giorgia and Kreiman, Gabriel},
  booktitle={International Conference on Machine Learning},
  pages={4937--4955},
  year={2022},
  organization={PMLR}
}

@inproceedings{ororbia2019biologically,
  title={Biologically motivated algorithms for propagating local target representations},
  author={Ororbia, Alexander G and Mali, Ankur},
  booktitle={Proceedings of the aaai conference on artificial intelligence},
  volume={33},
  number={01},
  pages={4651--4658},
  year={2019}
}

@inproceedings{lorberbom2024layer,
  title={Layer collaboration in the forward-forward algorithm},
  author={Lorberbom, Guy and Gat, Itai and Adi, Yossi and Schwing, Alexander and Hazan, Tamir},
  booktitle={Proceedings of the AAAI Conference on Artificial Intelligence},
  volume={38},
  number={13},
  pages={14141--14148},
  year={2024}
}

@inproceedings{giampaolo2023investigating,
  title={Investigating random variations of the forward-forward algorithm for training neural networks},
  author={Giampaolo, Fabio and Izzo, Stefano and Prezioso, Edoardo and Piccialli, Francesco},
  booktitle={2023 International Joint Conference on Neural Networks (IJCNN)},
  pages={1--7},
  year={2023},
  organization={IEEE}
}

@article{ororbia2023predictive,
  title={The predictive forward-forward algorithm},
  author={Ororbia, Alexander and Mali, Ankur},
  journal={arXiv preprint arXiv:2301.01452},
  year={2023}
}

@article{scodellaro2023training,
  title={Training convolutional neural networks with the forward-forward algorithm},
  author={Scodellaro, Riccardo and Kulkarni, Ajinkya and Alves, Frauke and Schr{\"o}ter, Matthias},
  journal={arXiv preprint arXiv:2312.14924},
  year={2023}
}

@inproceedings{reyes2024forward,
  title={Forward-forward algorithm for hyperspectral image classification},
  author={Reyes-Angulo, Abel A and Paheding, Sidike},
  booktitle={Proceedings of the IEEE/CVF Conference on Computer Vision and Pattern Recognition},
  pages={3153--3161},
  year={2024}
}

@article{miller1995wordnet,
  title={WordNet: a lexical database for English},
  author={Miller, George A},
  journal={Communications of the ACM},
  volume={38},
  number={11},
  pages={39--41},
  year={1995},
  publisher={ACM New York, NY, USA}
}

@inproceedings{deng2009imagenet,
  title={Imagenet: A large-scale hierarchical image database},
  author={Deng, Jia and Dong, Wei and Socher, Richard and Li, Li-Jia and Li, Kai and Fei-Fei, Li},
  booktitle={2009 IEEE conference on computer vision and pattern recognition},
  pages={248--255},
  year={2009},
  organization={Ieee}
}

@article{mikolov2013efficient,
  title={Efficient estimation of word representations in vector space},
  author={Mikolov, Tomas and Chen, Kai and Corrado, Greg and Dean, Jeffrey},
  journal={arXiv preprint arXiv:1301.3781},
  year={2013}
}

@article{xiao2017fashion,
  title={Fashion-mnist: a novel image dataset for benchmarking machine learning algorithms},
  author={Xiao, Han and Rasul, Kashif and Vollgraf, Roland},
  journal={arXiv preprint arXiv:1708.07747},
  year={2017}
}

@article{lecun2002gradient,
  title={Gradient-based learning applied to document recognition},
  author={LeCun, Yann and Bottou, L{\'e}on and Bengio, Yoshua and Haffner, Patrick},
  journal={Proceedings of the IEEE},
  volume={86},
  number={11},
  pages={2278--2324},
  year={2002},
  publisher={Ieee}
}

@inproceedings{coates2011analysis,
  title={An analysis of single-layer networks in unsupervised feature learning},
  author={Coates, Adam and Ng, Andrew and Lee, Honglak},
  booktitle={Proceedings of the fourteenth international conference on artificial intelligence and statistics},
  pages={215--223},
  year={2011},
  organization={JMLR Workshop and Conference Proceedings}
}

@article{maaten2008visualizing,
  title={Visualizing data using t-SNE},
  author={Maaten, Laurens van der and Hinton, Geoffrey},
  journal={Journal of machine learning research},
  volume={9},
  number={Nov},
  pages={2579--2605},
  year={2008}
}

@article{paliotta2023graph,
  title={Graph neural networks go forward-forward},
  author={Paliotta, Daniele and Alain, Mathieu and M{\'a}t{\'e}, B{\'a}lint and Fleuret, Fran{\c{c}}ois},
  journal={arXiv preprint arXiv:2302.05282},
  year={2023}
}

@inproceedings{jaderberg2017decoupled,
  title={Decoupled neural interfaces using synthetic gradients},
  author={Jaderberg, Max and Czarnecki, Wojciech Marian and Osindero, Simon and Vinyals, Oriol and Graves, Alex and Silver, David and Kavukcuoglu, Koray},
  booktitle={International conference on machine learning},
  pages={1627--1635},
  year={2017},
  organization={PMLR}
}

@article{lillicrap2020backpropagation,
  title={Backpropagation and the brain},
  author={Lillicrap, Timothy P and Santoro, Adam and Marris, Luke and Akerman, Colin J and Hinton, Geoffrey},
  journal={Nature Reviews Neuroscience},
  volume={21},
  number={6},
  pages={335--346},
  year={2020},
  publisher={Nature Publishing Group UK London}
}

@inproceedings{pau2023suitability,
  title={Suitability of forward-forward and pepita learning to mlcommons-tiny benchmarks},
  author={Pau, Danilo Pietro and Aymone, Fabrizio Maria},
  booktitle={2023 IEEE International Conference on Omni-layer Intelligent Systems (COINS)},
  pages={1--6},
  year={2023},
  organization={IEEE}
}

@inproceedings{baghersalimi2023layer,
  title={Layer-wise learning framework for efficient dnn deployment in biomedical wearable systems},
  author={Baghersalimi, Saleh and Amirshahi, Alireza and Teijeiro, Tomas and Aminifar, Amir and Atienza, David},
  booktitle={2023 IEEE 19th International Conference On Body Sensor Networks (BSN)},
  pages={1--4},
  year={2023},
  organization={IEEE}
}

@inproceedings{aghagolzadeh2024marginal,
  title={Marginal Contrastive Loss: A Step Forward for Forward-Forward},
  author={Aghagolzadeh, Hossein and Ezoji, Mehdi},
  booktitle={2024 13th Iranian/3rd International Machine Vision and Image Processing Conference (MVIP)},
  pages={1--6},
  year={2024},
  organization={IEEE}
}

@article{ahamed2023forward,
  title={Forward-forward contrastive learning},
  author={Ahamed, Md Atik and Chen, Jin and Imran, Abdullah-Al-Zubaer},
  journal={arXiv preprint arXiv:2305.02927},
  year={2023}
}

@inproceedings{wang2017normface,
  title={Normface: L2 hypersphere embedding for face verification},
  author={Wang, Feng and Xiang, Xiang and Cheng, Jian and Yuille, Alan Loddon},
  booktitle={Proceedings of the 25th ACM international conference on Multimedia},
  pages={1041--1049},
  year={2017}
}

@inproceedings{liu2017sphereface,
  title={Sphereface: Deep hypersphere embedding for face recognition},
  author={Liu, Weiyang and Wen, Yandong and Yu, Zhiding and Li, Ming and Raj, Bhiksha and Song, Le},
  booktitle={Proceedings of the IEEE conference on computer vision and pattern recognition},
  pages={212--220},
  year={2017}
}

@inproceedings{wang2018cosface,
  title={Cosface: Large margin cosine loss for deep face recognition},
  author={Wang, Hao and Wang, Yitong and Zhou, Zheng and Ji, Xing and Gong, Dihong and Zhou, Jingchao and Li, Zhifeng and Liu, Wei},
  booktitle={Proceedings of the IEEE conference on computer vision and pattern recognition},
  pages={5265--5274},
  year={2018}
}
